\DocumentMetadata{%
    pdfstandard=A-3u,
    pdfversion=1.7,
    lang=en-US,
}

\documentclass[nocopyright,varvw,hyphenate,pdf-a,subscriptcorrection,upint,mathalfa=cal=cm,balance]{asmejour}

\hypersetup{%
    pdfauthor = {Dasharadhan Mahalingam, Michael Gallagher, Nilanjan Chakraborty, Stanislaus S. Wong},
    pdftitle = {Robotic Nanoparticle Synthesis via Solution-based Processes},
    pdfkeywords = {Screw Geometry, Complex Manipulation Tasks, Constrained Motion Planning, Lab Automation},
    pdfsubject = {Robotics},
}

\JourName{Translational Robotics}

\usepackage{graphicx} 

\usepackage{array}

\usepackage{pgfplots}
\pgfplotsset{compat=1.18}

\begin{document}

\SetAuthorBlock{Dasharadhan Mahalingam\CorrespondingAuthor{}\EqualContribution}{
    Department of Mechanical Engineering,\\
    Stony Brook University,\\
    Stony Brook, NY 11794, USA\\
    email:\\
    \small{dasharadhan.mahalingam@stonybrook.edu}
}

\SetAuthorBlock{Michael Gallagher\EqualContribution}{
    Department of Chemistry,\\
    Stony Brook University,\\
    Stony Brook, NY 11794, USA\\
    email: michael.g.gallagher@stonybrook.edu
} 

\SetAuthorBlock{Nilanjan Chakraborty}{
    Department of Mechanical Engineering,\\
    Stony Brook University,\\
    Stony Brook, NY 11794, USA\\
    email: nilanjan.chakraborty@stonybrook.edu
} 

\SetAuthorBlock{Stanislaus S. Wong}{
    Department of Chemistry,\\
    Stony Brook University,\\
    Stony Brook, NY 11794, USA\\
    email: stanislaus.wong@stonybrook.edu
} 

\title{Robotic Nanoparticle Synthesis via Solution-based Processes}

\keywords{Screw Geometry, Complex Manipulation Tasks, Constrained Motion Planning, Lab Automation.}

\begin{abstract}
We present a screw geometry-based manipulation planning framework for the robotic automation of solution-based synthesis, exemplified through the preparation of gold and magnetite nanoparticles. The synthesis protocols are inherently long-horizon, multi-step tasks, requiring skills such as pick-and-place, pouring, turning a knob, and periodic visual inspection to detect reaction completion. A central challenge is that some skills, notably pouring, transferring containers with solutions, and turning a knob, impose geometric and kinematic constraints on the end-effector motion. To address this, we use a programming by demonstration paradigm where the constraints can be extracted from a single demonstration. This combination of screw-based motion representation and demonstration-driven specification enables domain experts, such as chemists, to readily adapt and reprogram the system for new experimental protocols and laboratory setups without requiring expertise in robotics or motion planning. We extract sequences of constant screws from demonstrations, which compactly encode the motion constraints while remaining coordinate-invariant. This representation enables robust generalization across variations in grasp placement and allows parameterized reuse of a skill learned from a single example. By composing these screw-parameterized primitives according to the synthesis protocol, the robot autonomously generates motion plans that execute the complete experiment over repeated runs. Our results highlight that screw-theoretic planning, combined with programming by demonstration, provides a rigorous and generalizable foundation for long-horizon laboratory automation, thereby enabling fundamental kinematics to have a translational impact on the use of robots in developing scalable solution-based synthesis protocols.

\end{abstract}

\maketitle

\begin{figure*}[ht!]
    \centering
    \begin{subfigure}[b]{0.3\textwidth}
        \includegraphics[width=\textwidth]{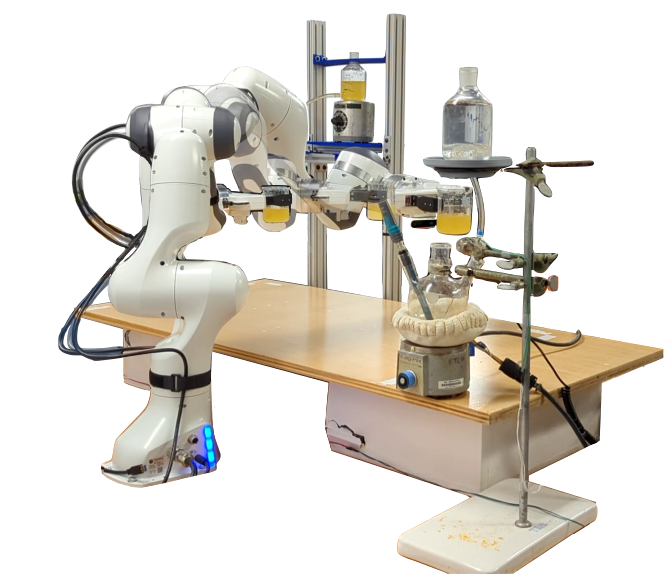}
        \caption{Transfer without spilling}
    \end{subfigure}
    \hfill
    \begin{subfigure}[b]{0.3\textwidth}
        \includegraphics[width=\textwidth]{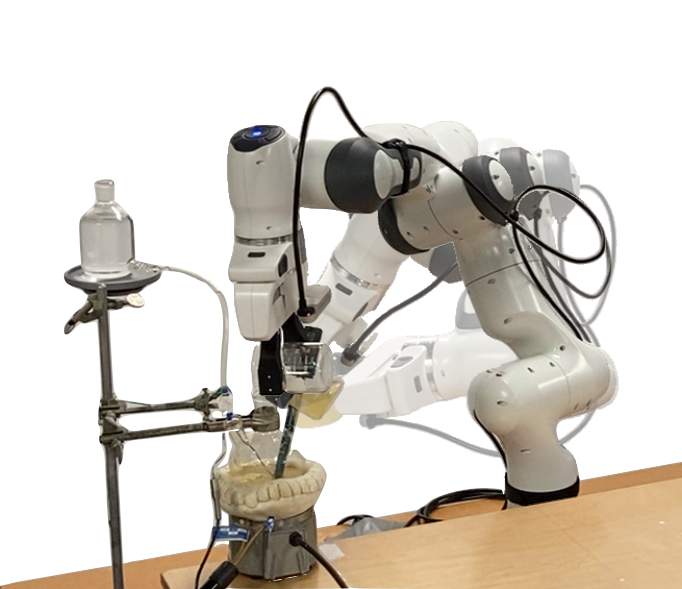}
        \caption{Pouring}
    \end{subfigure}
    \hfill
    \begin{subfigure}[b]{0.3\textwidth}
        \includegraphics[width=\textwidth]{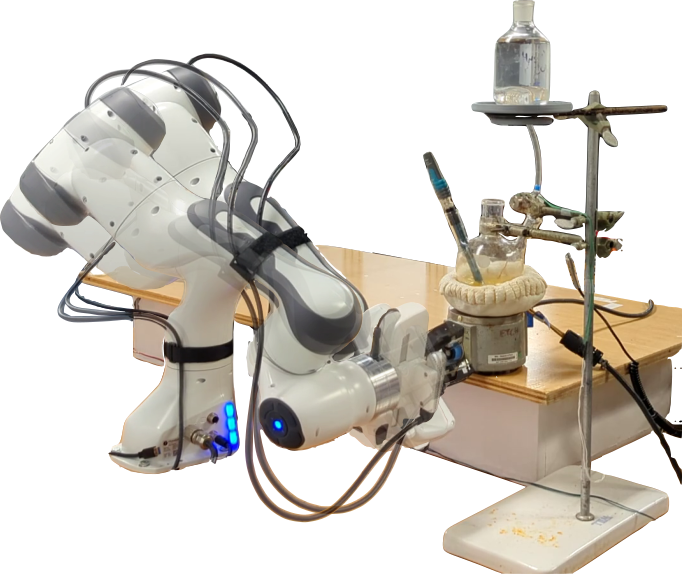}
        \caption{Operating Stir Plate}
    \end{subfigure}
    \caption{Robot performing constrained manipulation tasks as part of Magnetite Nano-Particle synthesis experiment.}
    \label{fig:main_fig}
\end{figure*}

\section{Introduction}

Automation in chemical laboratories has advanced significantly in recent years, particularly through the use of flow-reaction platforms that enable highly repeatable and continuous synthesis~\cite{vescovi2023towards}. While these systems provide a high degree of automation, they are limited in scope; many protocols in chemistry rely on solution-based processes that require physical manipulation of containers, liquids, and intermediate products. These protocols demand a robot to perform tasks involving constrained manipulation skills such as pouring, scooping, stirring, or carefully placing vessels under specific geometric and kinematic constraints. Despite the critical importance of such tasks, robotic systems that can robustly execute a variety of tasks that require constrained manipulation skills remain underdeveloped.

Traditional motion planning approaches have fundamental limitations in this context. In joint space–based methods~\cite{BerensonSK2014, JailletP2013, KingstonMK2018}, the constraints imposed by manipulation tasks (e.g., maintaining a fixed pouring axis) form nonlinear manifolds. These manifolds change with different grasp placements, since the representation is not coordinate-invariant. On the other hand, task space–based methods such as~\cite{Chiaverini2008} represent constraints directly in terms of end-effector position and orientation, but the resulting parameterizations are often nonlinear and similarly lack coordinate invariance. These properties make both approaches brittle to changes in grasp or task instance (placement of task-relevant objects), limiting their robustness in long-horizon protocols.

In this paper, we propose a screw geometry–based manipulation planning framework that addresses these shortcomings. Screw representations offer a coordinate-invariant and compact way to encode motion constraints, enabling robust generalization from a single demonstration to new task instances~\cite{mahalingam2023humanguided}. As shown in~\cite{mahalingam2023humanguided,mahalingam2024verticalfarming, laha2021userguided,laha2022userguided}, a multitude of manipulation tasks where the constraints arise from the manipulation of articulated structures (like doors and drawers) or from the nature of the task itself (like pouring and scooping) can be represented as a sequence of constant screws, which capture the essential kinematic structure of the task. The screw representation is coordinate-invariant, which implies that it is robust to choices of the coordinate frame on the object being manipulated or grasp placement variations. By extracting such screw-parameterized primitives from kinesthetic demonstrations, the robot acquires a library of reusable skills that can be flexibly composed to execute multistep laboratory protocols.
The parameterized skills enable the robot to plan motion for a new instance of the same task, allowing the robot to autonomously perform the full sequence of actions required for the synthesis. 

Given demonstrations of all the required tasks and knowledge of their ordering, along with visual cues to detect state changes and time delays where needed, the framework synthesizes joint space motion plans that the robot executes. We illustrate the approach based on the synthesis of two nano-scale motifs, namely gold and magnetite nanoparticles, generated via two distinct synthetic routes, as canonical examples of solution-based chemical synthesis wherein lessons learned can be reasonably applied to other analogous systems. Specifically, these examples include demonstrations of both room-temperature and heated solution-based methods and highlight the ability to control key reaction variables such as reagent concentrations autonomously. Although our experiments focus on this particular process, the framework is general and can be extended to other chemical syntheses by providing the appropriate demonstrations. Since many laboratory tasks share common primitives (e.g., pouring, pick-and-place), demonstrations can be reused across protocols, enabling the gradual development of a database of parameterized manipulation skills.

\noindent
{\bf Contributions}: The contributions of this paper are two-fold. First, we show that {a screw geometry–based programming by demonstration (PbD) approach~\cite{mahalingam2023humanguided} for teaching skills to robots} enables the execution of long-horizon, multistep protocols in solution-based synthesis, where the composition of multiple constrained manipulation skills must be robustly sequenced to achieve a complete experiment. Using gold nanoparticle and magnetite nanoparticle synthesis as representative case studies, we are able to show how demonstrations of individual primitives can be parameterized and reused to autonomously generate motion plans for the entire protocol. Second, we highlight the translational potential of this approach: by creating a reusable database of parameterized manipulation primitives, our framework moves toward flexible and reproducible laboratory automation that can extend the capabilities of human chemists, operate continuously, and accelerate discovery. While the present work focuses on nanoparticle synthesis, the framework generalizes naturally to other chemical protocols that require constrained manipulation, thereby positioning screw-theoretic planning as a foundation for scalable robotic chemists.

\noindent
{\bf Organization of the Paper}: The paper is organized as follows: In Section~\ref{sec:rw}, we present the related literature on the use of robotics in automating experimental protocols and also a brief review of the state-of-the-art in robotics on constrained manipulation planning. In Section~\ref{sec:problem}, we formalize our problem, and in Section~\ref{sec:approach}, we present our solution approach. The experimental results are presented and discussed in Section~\ref{sec:results} and the conclusions and future work are laid out in Section~\ref{sec:conc}. 

\section{Related Work}
\label{sec:rw}

\subsection{Laboratory Automation using Robots}

With the recent advances being made in the field of robotics, the use of general purpose robots to automate experiments in chemical laboratories has been gaining traction \cite{burger2020mobile, lim2021organic, darvish2024organa}. The use of such robots, while reducing the physical and mental fatigue experienced by chemists performing those reactions, also reduces the risks associated with the handling of harmful chemicals. Moreover, Self Driving Laboratories (SDL) consisting of automated systems which are capable of performing experiments and making decisions autonomously help to accelerate discovery of chemicals/materials \cite{tom2024selfdrivinglabs}. Such labs require automated systems that are not too specific to a particular reaction so that they can be used in a wide variety of reactions without much modification of existing laboratory setups. Hence, general purpose robots, which are capable of dexterous manipulation, are being used in lab automation to perform complex manipulation tasks and operate lab equipment that actual human chemists perform.

Researchers in \cite{burger2020mobile} have shown that robots can assist in experimental searches when the experimental complexity scales exponentially with the number of variables. They conducted a search for photo-catalysts that can be used for hydrogen production from water. \cite{ShiriLZGR+21} demonstrated an automated robotic platform for solubility screening. \cite{fakhruldeen2022archemist} proposes a standard architecture for integrating different robotic platforms and laboratory equipment for automating chemical experiments. \cite{pizzuto2024samplescraping} proposes a RL based approach to acquire the manipulation skills required for performing scraping. \cite{fakhruldeen2025behaviourtrees} highlights a method incorporating multi-modal perception within a behavior tree for verifying successful execution of a task. \cite{knobbe2022pipetting} demonstrated a robotic system capable of performing pipetting using force control. \cite{lim2021organic} describes a system capable of tasks such as performing pouring in addition to pick-and-place. \cite{darvish2024organa, yoshikawa2023llm} discusses a robotic assistant that utilizes LLMs to translate natural language into long-horizon task plans for execution using a constrained task and motion planner.

While previous research has shown the advantages of using robots to automate chemical experiments in laboratories, with respect to manipulation, their main focus was on developing systems that were capable of performing a finite set of manipulation tasks. In \cite{burger2020mobile}, pick-and-place was the only manipulation task considered, whereas \cite{ShiriLZGR+21} considered capping/uncapping procedures in addition to pick-and-place. Also, specialized controllers that are specific to certain tasks are proposed in \cite{knobbe2022pipetting, fakhruldeen2022archemist, darvish2024organa}. While \cite{pizzuto2024samplescraping} describes a learning approach for tasks, it is specific to the scraping task. Current research does not focus on how a wide variety of manipulation skills with complex constraints can be added to the repertoire of tasks that the robot can perform. By contrast, this work focuses on how people without prior robotics knowledge can teach robots a number of specific manipulation skills that may be necessary to perform a chemical reaction. The robot then uses these manipulation skills that were subsequently shown to autonomously run an experiment.

\subsection{Constrained Motion Planning}

This work features a framework that uses a single kinesthetic demonstration provided by the user to perform a manipulation task. Manipulation tasks performed in chemical synthesis may have complex constraints on the object motion that need to be satisfied. For example, consider tasks such as pouring liquids from one container into another and, transferring containers filled with liquids without spilling. These constraints may be hard to describe as they may form nonlinear manifolds in the task-space as well as the joint-space. Representing such constraints analytically is non-trivial, and many Learning from Demonstration (LfD) \cite{billard2016learning} approaches have been proposed to facilitate their autonomous execution on robots. Motion primitive based approaches \cite{gutierrez2025movement} based on GMMs \cite{calinon2007, calinon2014tpgmm} and HMMs \cite{calinon2010, calinon2011, niekum2013incremental} require multiple demonstrations. DMP \cite{IjspeertNHPS13} based approaches use a single demonstration but require tuning of the hyper-parameters.

Human provided tele-operated demonstrations have been used along with diffusion-based \cite{bekris2025diffusion} and transformer-based \cite{zhao2023learning} approaches for learning tasks. However, such approaches require significantly more demonstrations \cite{oneill2024openx} and places the burden on the end user to provide multiple demonstrations for every task.

Prior work on using the screw geometry-based representation of motion \cite{mahalingam2023humanguided, mahalingam2024verticalfarming} has shown that it can be used to successfully extract task constraints from a single kinesthetic demonstration for autonomous execution. This work highlights that it can be extended to create a database of reusable manipulation primitives, which can then be sequenced together to perform the manipulation tasks involved in an experiment.

\section{Mathematical Preliminaries}
\label{sec:math_prelims}

In this section, we provide a brief review of the background knowledge required for this work. The joint space of the robot is the set of all possible joint configurations and is denoted by $\mathcal{J} \subset \mathbb{R}^n$ where $n$ is the number of degrees of freedom (DoF) of the robot. The set of all rigid body configurations is $SE(3)$, the Special Euclidean Group of dimension $3$. An element of $SE(3)$ is also referred to as a \textit{pose}. Thus, the task space $\mathcal{T}$ of the robot, i.e., the set of end-effector poses, is a subset of $SE(3)$. The joint configuration of the robot, $\mathbf{\Theta} \in \mathcal{J}$, is a vector of length $n$. For every valid joint configuration $\mathbf{\Theta}$, the forward kinematics of the manipulator maps it to a unique end-effector pose, $\mathbf{G} \in \mathcal{T}$. 

The synthesis of a chemical compound involves a sequence of manipulation tasks. For example, consider the following sequence of tasks: pour solution A from a test-tube into a beaker containing solution B, then add salt C into the beaker, and finally stir the solution. The above synthesis process involves three distinctive and different manipulation tasks. Throughout this paper, when we refer to a task, we mean a specific manipulation task involved in the synthesis of a compound.

\subsection{Task Instance}
Objects whose poses affect the generation of manipulation plans for performing a task are called $\textit{task-relevant objects}$. A $\textit{task instance}$ is defined as the set of all task-relevant object poses, and it is denoted by $\mathcal{O} = \{\mathbf{O}_1, \mathbf{O}_2, \dots, \mathbf{O}_m\}$. For the simple task mentioned earlier in Section \ref{sec:math_prelims}, the task instance consists of the poses of the test-tube containing solution A, the beaker containing solution B, and the container containing salt C, $\mathcal{O} = \{\mathbf{O}_A, \mathbf{O}_B, \mathbf{O}_C\}$.

\subsection{Kinesthetic Demonstration}
A user can provide a kinesthetic demonstration $\mathcal{D}$ by enabling the zero gravity mode and then moving the robot's end-effector to perform the manipulation task. A particular demonstration $\mathcal{D}$ is associated with a task instance $\mathcal{O}$. The kinesthetic demonstration is recorded as a sequence of joint angles, i.e., a path in $\mathcal{J}$, which can be mapped to a path in the task space $\mathcal{T}$ using the forward kinematics map. Throughout this paper, whenever we mention a demonstration, we refer to the sequence of poses, $\mathcal{G} = \langle\mathbf{G}_1, \mathbf{G}_2, \mathbf{G}_3, ...\rangle$ in the task space $\mathcal{T}$ that the end-effector of the robot goes through, while performing the task along with its associated task instance, $\mathcal{D} = (\mathcal{G}, \mathcal{O})$.

\subsection{Screw Displacement}
The Chasles-Mozzi theorem states that the general Euclidean displacement/motion of a rigid body from the origin $\mathbf{I}$ to $\mathbf{T} = (\mathbf{R},\boldsymbol{p}) \in SE(3)$
can be expressed as a rotation $\theta$ about a fixed axis $\mathcal{S}$, called the \textit{screw axis}, and a translation $d$ along that axis. 
Plücker coordinates can be used to represent the screw axis by $\boldsymbol{\omega}$ and $\boldsymbol{m}$, where $\boldsymbol{\omega} \in \mathbb{R}^3$ is a unit vector that represents the direction of the screw axis, $\boldsymbol{m} = \boldsymbol{r} \times \boldsymbol{\omega}$, and $\boldsymbol{r} \in \mathbb{R}^3$ is an arbitrary point on the screw axis. Thus, the screw parameters are defined as $\boldsymbol{\omega}, \boldsymbol{m}, h, \theta$, where $h$ is the pitch of the screw and $\theta$ is its magnitude. In general, for pure rotation and general screw motion, $h$ is finite, while for pure translation, $h = \infty$.
If $\mathbf{R} \neq \mathbf{I}$, then by using the standard procedure to obtain rotation axis and magnitude from the rotation matrix $\mathbf{R}$, we can determine $\bm{\omega}$ and $\theta$. The pitch is given by $h = \bm{\omega}^T\bm{\upsilon}$ and $\mathbf{m} = \bm{\upsilon} - h\bm{\omega}$ where $\bm{\upsilon} = \left[(\mathbf{I} - e^{\hat{\bm{\omega}}\theta})\hat{\bm{\omega}} + \theta\bm{\omega}\bm{\omega}^T\right]^{-1}\bm{p}$. If $\mathbf{R} = \mathbf{I}$, then the motion is pure translation where $h = \infty$ and $\bm{m} = \bm{0}$ by definition. We can obtain $\theta$ and $\bm\omega$ from $\theta = ||\bm{p}||$ and $\bm{\omega} = \bm{p}/||\bm{p}||$.
\textbf{A {\em constant screw motion} is a motion where the parameters $\boldsymbol{\omega}, \boldsymbol{m}$, and $h$ stay constant throughout the motion}.

Given the screw parameters $\boldsymbol{\omega}, \boldsymbol{m}$ and $h$, the screw displacement for a motion of magnitude $\theta$ can be obtained using the matrix exponential, $\mathbf{T} = e^{\widehat{\bm\xi}\theta}$. Here, $\widehat{\bm\xi} \in se(3)$ and $\bm\xi \in \mathbb{R}^6$ are the unit twist and unit twist coordinates associated with the motion. They are defined as,
\begin{gather}
    \label{eq:unit_twist_general_screw}
    \widehat{\bm{\xi}} =
    \begin{bmatrix}
        \widehat{\bm{\omega}} & \bm{m} + h\bm{\omega} \\ 0 & 0
    \end{bmatrix},~ \bm{\xi} = 
    \begin{bmatrix}
        \bm{m} + h\bm{\omega} \\ \bm{\omega}
    \end{bmatrix}
    ~\text{for}~h\neq\infty \\
    \label{eq:unit_twist_prismatic}
    \widehat{\bm{\xi}} =
    \begin{bmatrix}
        \mathbf{I} & \bm{\omega} \\ 0 & 0
    \end{bmatrix},~ \bm{\xi} = 
    \begin{bmatrix}
        \bm{\omega} \\ \bm{0}
    \end{bmatrix}
    ~\text{for}~h=\infty
\end{gather}
It is noteworthy that while the matrix exponential maps elements of $se(3)$ to $SE(3)$, the corresponding matrix logarithm maps elements of $SE(3)$ to $se(3)$. Please refer to \cite{lynch2017modern, murray2017mathematical} for more details.
\begin{gather}
    e^{\widehat{\bm\xi}\theta} = \mathbf{T},
    \quad
    \log{\mathbf{T}} = \widehat{\bm\xi}\theta
\end{gather}

\subsection{Screw Linear Interpolation (ScLERP)}
To perform a one DoF smooth screw motion (with a constant rotation and translation rate) between two poses in $SE(3)$, Screw Linear interpolation (ScLERP) can be used. ScLERP generates a geodesic motion between two given poses $\mathbf{G}_i, \mathbf{G}_f \in SE(3)$. ScLERP can be performed using,
\begin{gather}
    \label{eq:sclerp}
    \mathbf{G}_\tau = e^{\widehat{\bm\xi}\tau\theta} \mathbf{G}_i
\end{gather}
where $\widehat{\bm{\xi}}\theta = \log(\mathbf{G}_f {\mathbf{G}^{-1}_i})$ is the twist that results in the screw displacement form $\mathbf{G}_i$ to $\mathbf{G}_f$ and $\tau\in[0,1]$ is the linear interpolation factor.

\section{Problem Formulation}
\label{sec:problem}
Consider the action of a robot performing chemical synthesis. The action can be decomposed as a sequence of tasks with the end goal being defined as the successful execution of all the individual tasks in sequence.
While successful motion plan generation and execution are a required condition for a task to be successfully performed, there may be additional conditions that need to be satisfied, depending upon the nature of the task, and come from the domain knowledge of the user. In this paper, we consider three other conditions: 1) Waiting for ``$t$'' amount of time before/after motion execution, 2) Monitoring for visual color changes of the products of chemical synthesis, and 3) Monitoring the pH values of the solutions.

\begin{figure*}[ht!]
    \centering
    \setlength{\tabcolsep}{3pt}
    
    \begin{tabular}{c c c c c}

        \rotatebox[origin=c]{90}{Pouring} &
        \begin{subfigure}[c]{0.23\textwidth}
            \includegraphics[width=\linewidth]{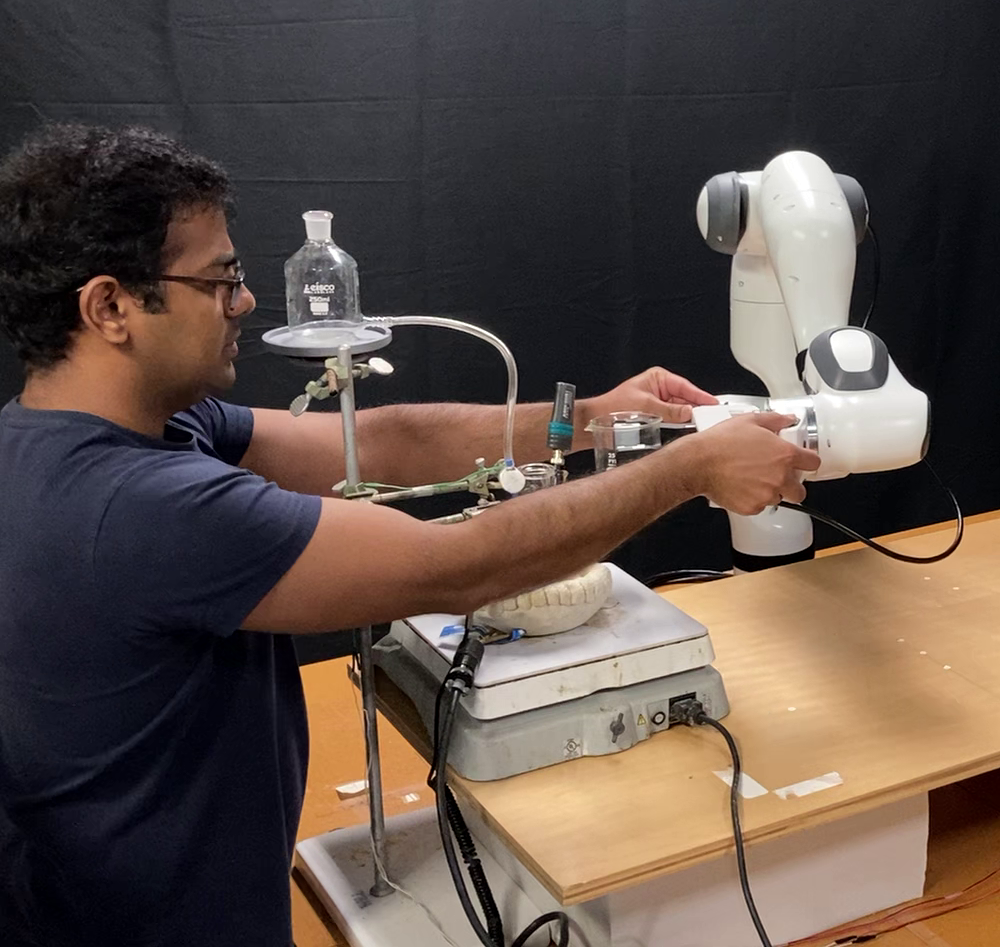}
        \end{subfigure} &
        \begin{subfigure}[c]{0.23\textwidth}
            \includegraphics[width=\linewidth]{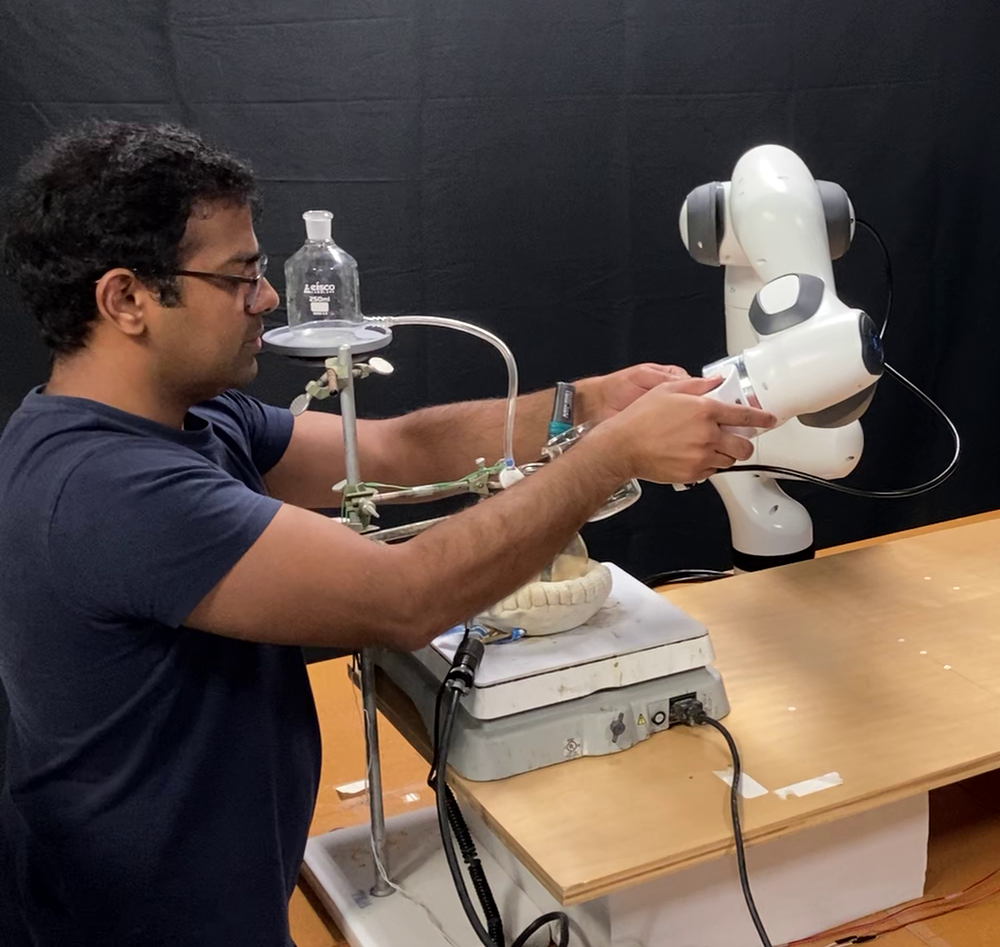}
        \end{subfigure} &
        \begin{subfigure}[c]{0.23\textwidth}
            \includegraphics[width=\linewidth]{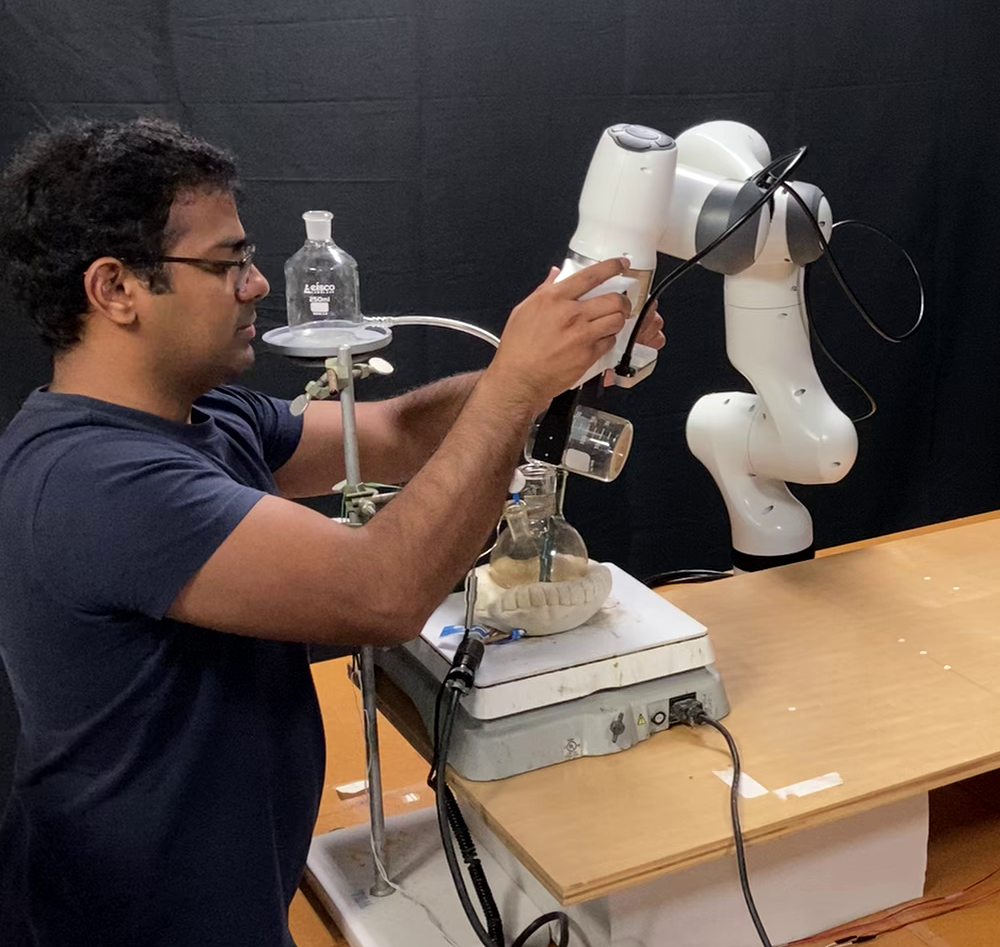}
        \end{subfigure} &
        \begin{subfigure}[c]{0.23\textwidth}
            \includegraphics[width=\linewidth]{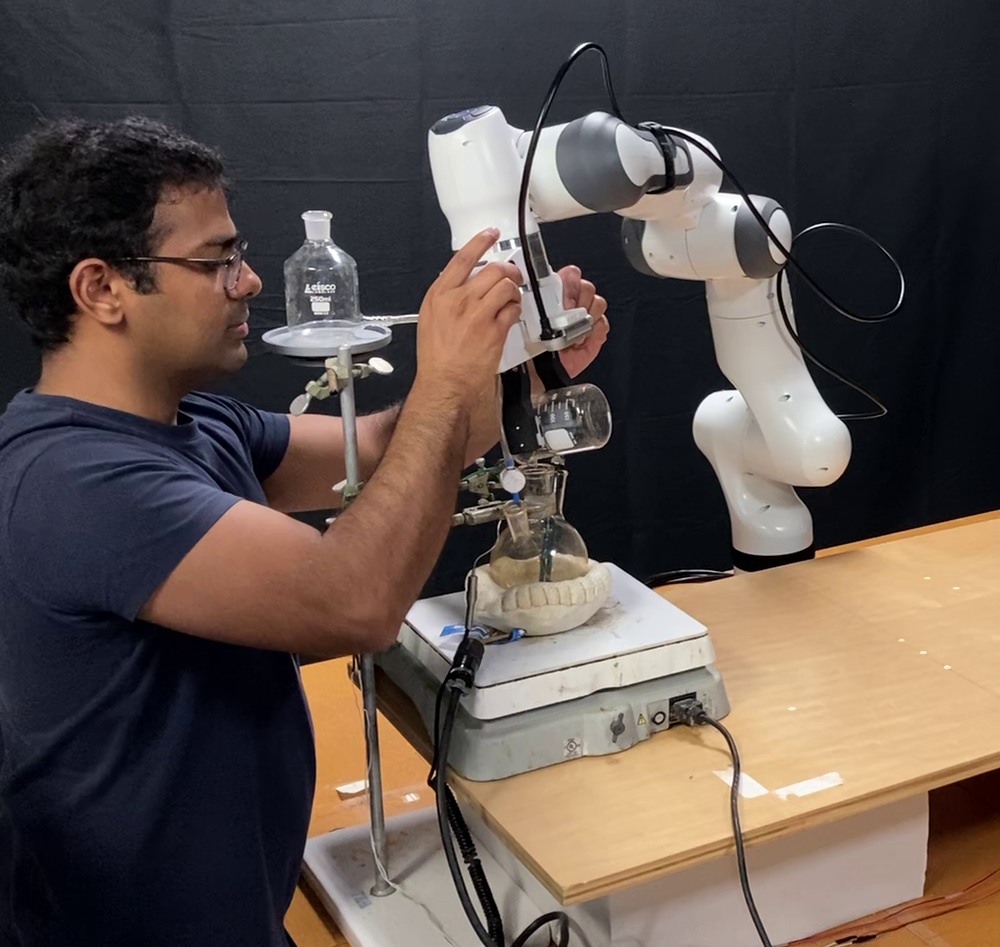}
        \end{subfigure} \\

        \noalign{\vspace{6pt}}

        \rotatebox[origin=c]{90}{Operating Stir Plate} &
        \begin{subfigure}[c]{0.23\textwidth}
            \includegraphics[width=\linewidth]{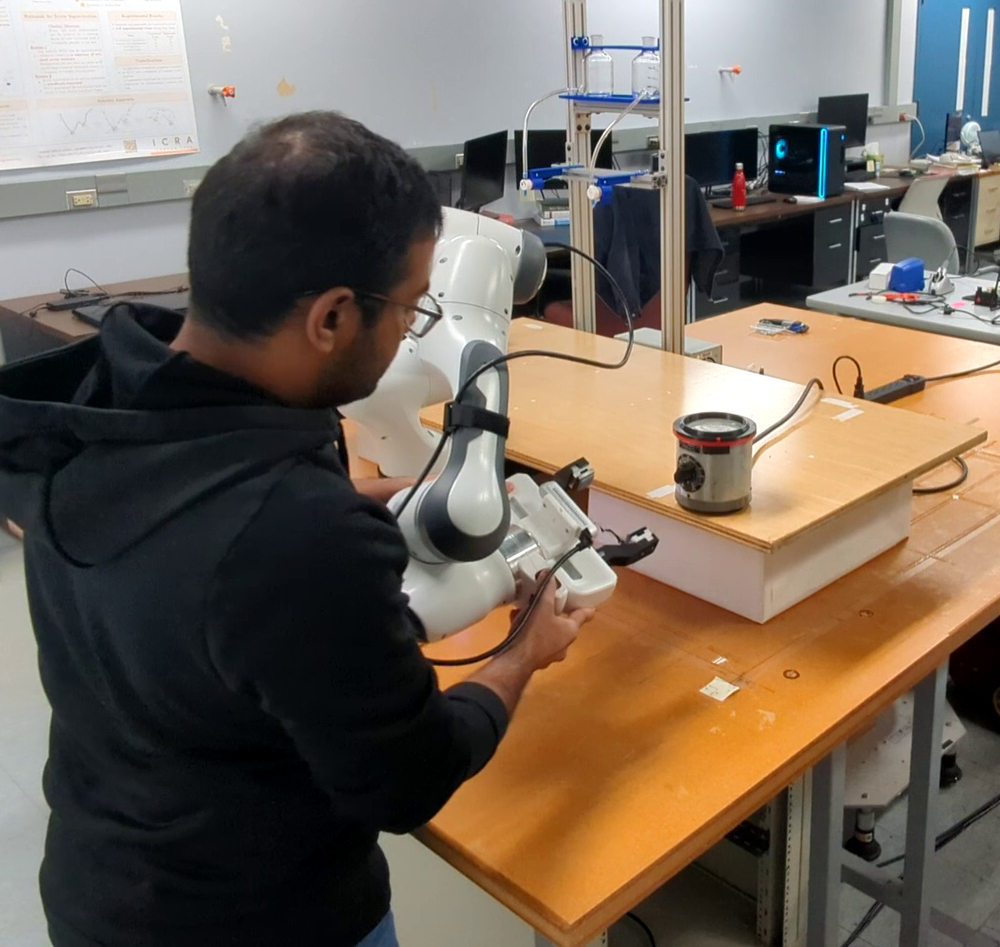}
        \end{subfigure} &
        \begin{subfigure}[c]{0.23\textwidth}
            \includegraphics[width=\linewidth]{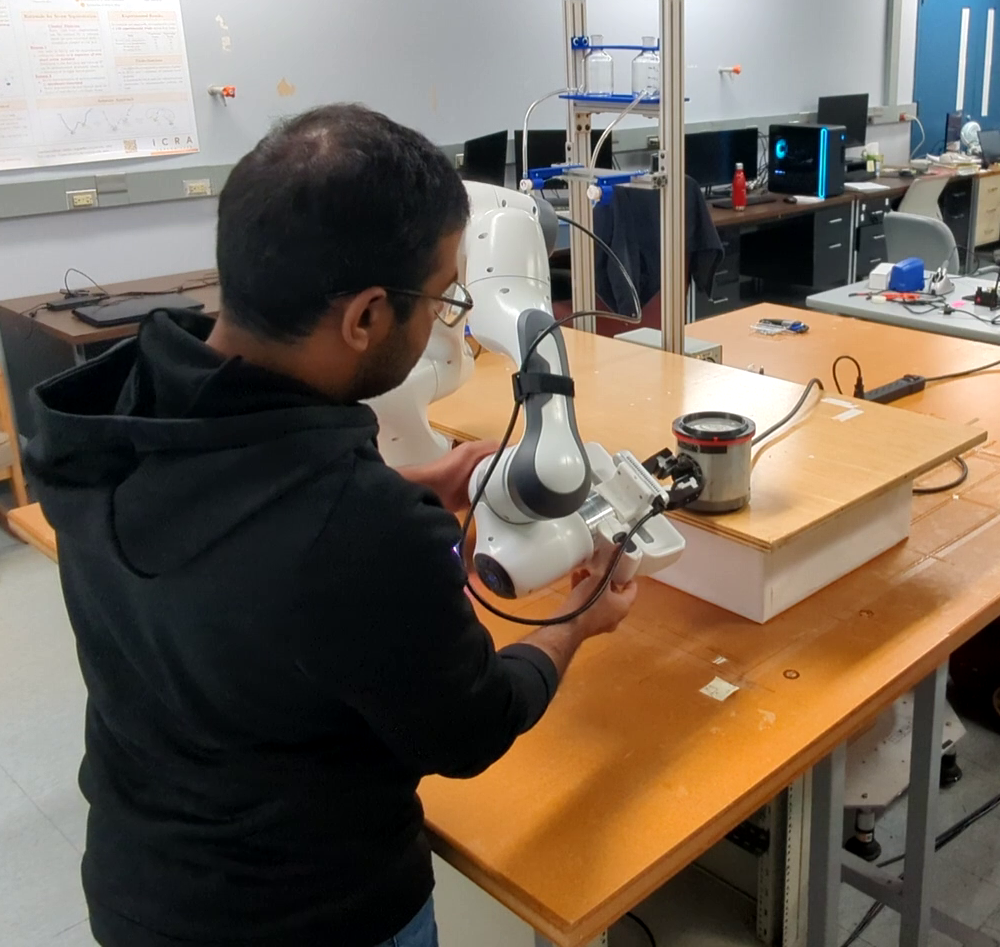}
        \end{subfigure} &
        \begin{subfigure}[c]{0.23\textwidth}
            \includegraphics[width=\linewidth]{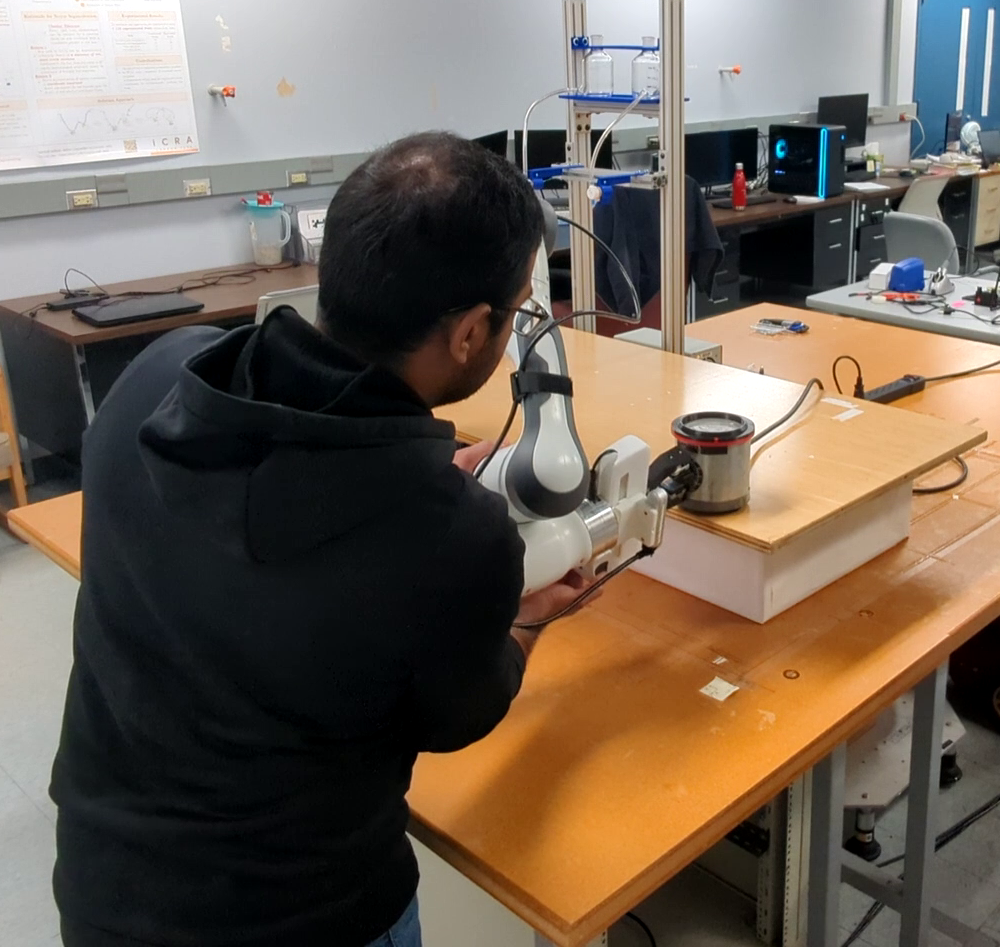}
        \end{subfigure} &
        \begin{subfigure}[c]{0.23\textwidth}
            \includegraphics[width=\linewidth]{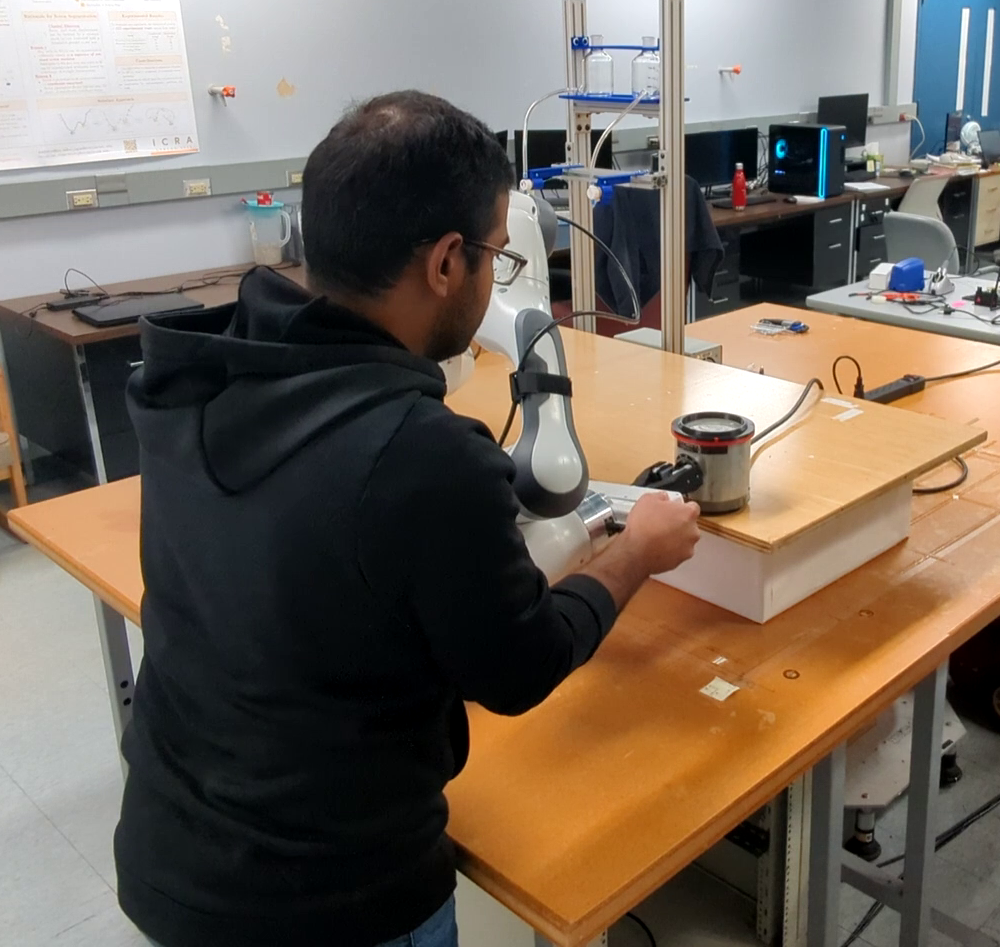}
        \end{subfigure}%
        
    \end{tabular}
    
    \caption{User provided kinesthetic demonstrations for the manipulation tasks involved in the nanoparticle synthesis.}
    \label{fig:kinesthetic_demo}
\end{figure*}

Let us define a mapping $f\left(\mathcal{L}\right): \mathcal{L} \rightarrow \mathcal{D}$ which maps the manipulation task labels to its appropriate demonstration. Here $\mathcal{L} = \{\texttt{pour}, \texttt{stir}, \texttt{pick}, \texttt{place}, \dots\}$ is described as the set of manipulation task labels and $\mathcal{D} = \{\mathcal{D}_{\texttt{pour}}, \mathcal{D}_{\texttt{stir}}, \mathcal{D}_{\texttt{pick}}, \mathcal{D}_{\texttt{place}}, \dots\}$ is defined as the set of demonstrations for each of the manipulation tasks involved in the chemical compound synthesis. Please note that we are overloading the notation $\mathcal{D}$ to represent a set of demonstrations and the sub-scripted notation $\mathcal{D}_{\texttt{x}}$ to represent a single demonstration for the task ``$\texttt{x}$''.

Given the mapping $f\left(\mathcal{L}\right): \mathcal{L} \rightarrow \mathcal{D}$, along with the sequence of manipulation tasks $\mathcal{M} = \langle (\texttt{pick}, \mathcal{O}'_{\texttt{pick}}), (\texttt{pour}, \mathcal{O}'_{\texttt{pour}}), \dots \rangle$, which defines the order in which the manipulation tasks have to be carried out with their corresponding task instance, the problem of chemical compound synthesis can be stated as: \textit{\textbf{Determine the sequence of joint configurations $\mathbf{\Theta} = \langle \bm{\theta}_1, \bm{\theta}_2, \dots \rangle, \bm{\theta}_i \in \mathbb{R}^n$, that would allow the robot to perform the given sequence of manipulation tasks $\mathcal{M}$, while satisfying the task constraints to successfully synthesize the chemical compound under consideration}}.

Here, the sets $\mathcal{L}$, $\mathcal{D}$ and the sequence $\mathcal{M}$ are user-defined depending upon the processes involved in the synthesis of the chemical compound under consideration. The user is required to provide a single kinesthetic demonstration $\mathcal{D}_{\texttt{x}}$ for every manipulation task ``$\texttt{x}$'', that has to be performed for the synthesis. Figure \ref{fig:kinesthetic_demo} shows the user providing a kinesthetic demonstration for the pouring and operating stir plate tasks that need to be performed in the magnetite nanoparticle synthesis experiment.

\section{Solution Approach}
\label{sec:approach}
In this section, we will go over the solution approach we use to automate the synthesis of nanoparticles using robots.

\subsection{User Guided Motion Planning}

This work uses a screw-geometry based representation of the motion involved in the manipulation tasks \cite{mahalingam2023humanguided}. We represent these task constraints as a sequence of \textbf{\textit{constant screw segments}}. The rationale behind the screw segmentation of motion is implied by the Chasles theorem: {\em Any curve in $SE(3)$, which represents a rigid body motion, can be approximated arbitrarily closely by a sequence of constant screw motions (or one-parameter subgroups of $SE(3)$)}. This is analogous to approximating a curve in space as a sequence of straight line segments. The screw-geometry based representation of the motion constraints is coordinate invariant (i.e. it does not depend on the choice of global or body reference frame assignment) allowing us to successfully transfer these constraints to a new task instance.

We first collect a single kinesthetic demonstration of every manipulation task involved in the synthesis. These demonstrations implicitly embed the constraints on the motion of the object that need to be satisfied in order to perform the task successfully. Consider a demonstration of task ``\texttt{x}'', denoted as $\mathcal{D}_{\texttt{x}} = \left(\mathcal{G}_{\texttt{x}}, \mathcal{O}_{\texttt{x}}\right)$, given by the user. The sequence of $SE(3)$ poses, $\mathcal{G}_{\texttt{x}} = \langle \mathbf{G}_{\texttt{x},1}, \mathbf{G}_{\texttt{x},2}, \dots, \mathbf{G}_{\texttt{x},m} \rangle$, describes the motion that the end-effector of the robot goes through to successfully perform task ``\texttt{x}''. For the robot to successfully execute the task, given a new task instance $\mathcal{O}'_{\texttt{x}}$, which describes the new pose of the task related object, it should be ensured that the task constraints embedded in the sequence $\mathcal{G}_{\texttt{x}}$ for the task instance $\mathcal{O}_{\texttt{x}}$ are extracted and transferred to the new task instance $\mathcal{O}'_{\texttt{x}}$.

\begin{figure*}[ht!]
    \begin{subfigure}[b]{\textwidth}
        \includegraphics[width=\textwidth]{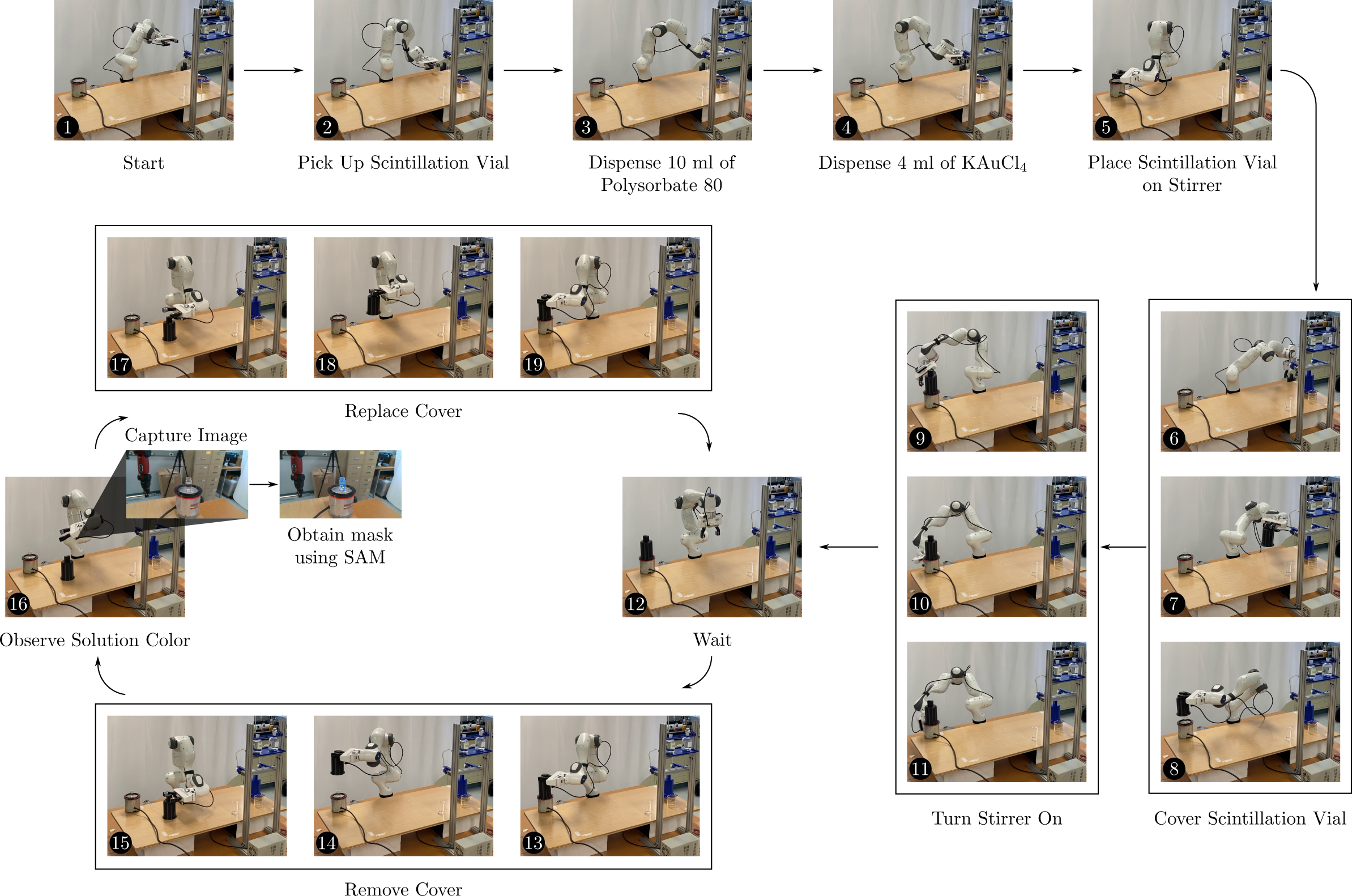}
    \end{subfigure}
    \caption{Automated Gold Nanoparticle Synthesis using Robots: Sequence of tasks involved in the synthesis of Gold Nanoparticles.}
    \label{fig:gold_np_exp}
\end{figure*}

To determine these task constraints, we segment the demonstrated motion $\mathcal{G}_{\texttt{x}}$ as a sequence of constant screw segments, $\mathcal{G}_{\texttt{x}, seg} = \langle \mathbf{G}_{\texttt{x},1}, \mathbf{G}_{\texttt{x},k_1}, \mathbf{G}_{\texttt{x},k_2} \dots, \mathbf{G}_{\texttt{x},k_u}, \mathbf{G}_{\texttt{x},m}\rangle, 2 \leq k_i < m,~\forall~i = 1,2,\dots,u$. The segmented motion $\mathcal{G}_{\texttt{x}, seg}$ is a sub-sequence of the demonstrated motion $\mathcal{G}_{\texttt{x}}$ such that each pair of consecutive poses in the segmented motion, $(\mathbf{G}_{\texttt{x},j}, \mathbf{G}_{\texttt{x},{j+1}}),~\text{where,}~j = 1, k_1, k_2, ..., k_u$, defines a constant screw segment with the entire demonstration consisting of $u+1$ constant screw segments. By following the heuristic, that the essential task constraints are observed in the demonstrated motion within a region surrounding the task relevant objects, we identify the screw segments in the sequence $\mathcal{G}_{\texttt{x}, seg}$ that lie within this region-of-interest surrounding the pose of the task related objects $\mathcal{O}_{\texttt{x}}$ during the demonstration. In this work, we define the region-of-interest to be a sphere of radius $r$ meters with its center located at the position of the task related object. If there are a total of $v$ task related objects, then we construct the region-of-interest for each of the $v$ objects in $\mathcal{O}_{\texttt{x}}$ and identify the screw segments associated with each object $i, 1 \leq i \leq v$ as,
$\langle \mathbf{G}_{\texttt{x}, l_{i,1}}, \mathbf{G}_{\texttt{x}, l_{i,2}}, \dots, \mathbf{G}_{\texttt{x}, l_{i,u_i}} \rangle$ where $l_{i,1}, l_{i,2}, \dots, l_{i,u_i}$ represent the indices of the $SE(3)$ pose in the sequence $\mathcal{G}_{\texttt{x}, seg}$ that lie within the region-of-interest of object $i$. Each object $i$ has a total of $u_i$ screw segments associated with it. The screw constraints corresponding to $\mathbf{O}_i$ will be referred to as the \textit{guiding poses} associated with object $i$, $\mathcal{GP}_{\texttt{x}, i} = \langle \mathbf{O}_i^{-1}\mathbf{G}_{\texttt{x}, l_{i,1}}, \mathbf{O}_i^{-1}\mathbf{G}_{\texttt{x}, l_{i,2}}, \dots, \mathbf{O}_i^{-1}\mathbf{G}_{\texttt{x}, l_{i,u_i}} \rangle$. The guiding poses associated with the task $\texttt{x}$ can then be expressed as the sequence of guiding poses associated with all of the $v$ objects, $\mathcal{GP}_{\texttt{x}} = \langle \mathcal{GP}_{\texttt{x},1}, \mathcal{GP}_{\texttt{x},2}, \dots, \mathcal{GP}_{\texttt{x},v} \rangle$. These guiding poses represent the sequence of constant screw motions that the end-effector of the robot goes through relative to each task- relevant object to successfully perform the task. Given a new task instance, $\mathcal{O}' = \{\mathbf{O}_1', \mathbf{O}_2', \dots, \mathbf{O}_v'\}$, the guiding poses associated with each object $i$ can be recomputed as, $\mathcal{GP}_{\texttt{x}, i}' = \langle \mathbf{O}_i'\mathbf{O}_i^{-1}\mathbf{G}_{\texttt{x}, l_{i,1}}, \mathbf{O}_i'\mathbf{O}_i^{-1}\mathbf{G}_{\texttt{x}, l_{i,2}}, \dots, \mathbf{O}_i'\mathbf{O}_i^{-1}\mathbf{G}_{\texttt{x}, l_{i,u_i}} \rangle$, giving us the guiding poses associated with the new task instance to be $\mathcal{GP}_{\texttt{x}}' = \langle \mathcal{GP}_{\texttt{x},1}', \mathcal{GP}_{\texttt{x},2}', \dots, \mathcal{GP}_{\texttt{x},v}' \rangle$. The recomputed guiding poses give us the sequence of constant screw motions that the end-effector of the robot needs to go through to successfully perform the task for the new task instance $\mathcal{O}'$.

{For more details on the representation of motion constraints as a sequence of constant screw segments and the segmentation algorithm used to extract the constant screws, please refer to \cite{mahalingam2023humanguided}.}

Once the guiding poses associated with the new task instance are computed, we can then plan for the joint space path of the robot to perform the task. We need to ensure that while computing the joint space path, the motion of the robot's end-effector satisfies the constant screw constraints defined by each pair of consecutive guiding poses. While the consecutive poses in the sequence $\mathcal{GP}_{\texttt{x},i}'$ define a constant screw constraint associated with object $i$, when going in-between the poses associated with objects $i$ and $i+1$ (when going from the last pose in $\mathcal{GP}_{\texttt{x},i}'$ to the first pose in $\mathcal{GP}_{\texttt{x},i+1}'$), we enforce the constant screw constraint as this ensures that for transfer tasks, the orientation of the container is maintained.

To determine the joint space path $\mathbf{\Theta}$, that ensures the constant screw constraints on the end-effector motion, we use a ScLERP based planner \cite{sarker2020sclerp} combined with Resolved Motion Rate Control (RMRC) \cite{Whitney1969ResolvedMotion}. Using ScLERP ensures that the computed joint space path satisfies the constant screw constraints.

\subsection{Chemical Synthesis using Robots}

Using the User Guided Motion Planning approach, given a set of demonstrations $\mathcal{D} = \{\mathcal{D}_{\texttt{pour}}, \mathcal{D}_{\texttt{stir}}, \mathcal{D}_{\texttt{pick}}, \mathcal{D}_{\texttt{place}}, \dots\}$, we can determine the guiding poses for each of the tasks relative to the task related objects as, $\{\mathcal{GP}_{\texttt{pour}}, \mathcal{GP}_{\texttt{stir}}, \mathcal{GP}_{\texttt{pick}}, \mathcal{GP}_{\texttt{place}}, \dots \}$. By determining the guiding poses for all the given demonstrations, we can compute the joint space path required to perform any of the task for which we have provided a demonstration if we are given a new task instance. Consider that the synthesis of a chemical compound requires a sequence of the following manipulation tasks in order: $\{ \texttt{pick}, \texttt{pour}, \texttt{place}, \texttt{stir} \}$. Assuming that the task instances for each of the above tasks, containing the poses of the relevant objects are $\{ \mathcal{O}'_{\texttt{pick}}, \mathcal{O}'_{\texttt{pour}}, \mathcal{O}'_{\texttt{place}}, \mathcal{O}'_{\texttt{stir}}\}$, we can construct the sequence of manipulation tasks, $\mathcal{M} = \langle (\texttt{pick}, \mathcal{O}'_{\texttt{pick}}), (\texttt{pour}, \mathcal{O}'_{\texttt{pour}}),  (\texttt{place}, \mathcal{O}'_{\texttt{place}}), (\texttt{stir}, \mathcal{O}'_{\texttt{stir}})\rangle$, and then determine the joint space path required to perform the tasks in sequence while satisfying the task constraints by following the proposed approach. In this manner, if the user collects the demonstration set $\mathcal{D}$ consisting of a single demonstration for each of the manipulation task involved and constructs the sequence of manipulation tasks, $\mathcal{M}$ for the synthesis of the required chemical compound, we can determine the required motion plan, $\mathbf{\Theta}$.

{In this work, we assume that the user does the necessary experimental setup in terms of placing the required hardware/glassware, provides the kinesthetic demonstration for all manipulation tasks and constructs the sequence of manipulation tasks, $\mathcal{M}$, involved in the experiment.}

\section{Experimental Results}

\label{sec:results}
{To evaluate our approach, we first validate the User Guided Motion Planner to show that the screw-based representation of motion is an effective method for representing motion constraints and allows us to transfer them to new task instances. We then conducted two case studies on the synthesis of nanoparticles using a robot.}
We considered the synthesis of 1) Gold (Au) Nanoparticles (NPs) and 2) Magnetite Nanoparticles.

\subsection{Experimental Hardware Setup}
Our experimental setup consisted of a 7 DoF Franka Emika Panda robot with the necessary glassware and other chemical equipment (solution dispenser, stir plate, temperature controller, etc.) placed within the workspace of the robot. The robot was equipped with a Realsense D415 RGBD camera in an eye-in-hand configuration. We used the camera to observe visual changes in the chemical reaction, such as color changes. We do not use the camera to perceive the environment and localize task related objects. We assume that the robot knows where the objects and other equipment are located. Since we are operating in a chemical laboratory environment, it is a reasonable assumption that objects are placed in an orderly manner at specific locations around the lab. Apart from the robot performing demonstrated manipulation tasks and moving the end-effector to bring the relevant containers within the field of view of the camera for perception, we have the following automation systems interfaced with the robot: 1) Solenoid valves connected to a reservoir for dispensing precise volumes of solutions, 2) pH probe for measuring the solution pH, and 3) Relays to turn on/off the temperature controller for heating solutions. These systems are used in conjunction with the manipulation tasks to synthesize the chemicals. An Arduino UNO R4 WiFi micro-controller connected to a L293D Driver shield was used for controlling the solenoids and relays. The solenoids were gravity fed from a reservoir containing the required solutions for the chemical synthesis. For pH measurements, we used a Vernier brand Go Direct pH Sensor connected via Bluetooth. As these tasks are not associated with manipulation, we will not go into the details of these systems.

For all the conducted experiments, we used the following parameters for the screw segmentation, $\epsilon_p = 0.01~\text{meter, and } \epsilon_\phi = 0.15$. These values for the screw segmentation were determined based on the empirical evidence reported in \cite{mahalingam2023humanguided}. We also set the radius of region-of-interest to be $0.15$ meters.

\subsection{Validation of the User Guided Planner}

{In order to validate the User Guided Planner, we collected a single demonstration for each of the manipulation tasks, $\{\texttt{pick}, \texttt{pour}, \texttt{place}\}$ and used the planner to plan the motion for new task instances of those tasks and executed it on the robot. Using the collected demonstrations, $\mathcal{D} = \{\mathcal{D}_{\texttt{pick}}, \mathcal{D}_{\texttt{pour}}, \mathcal{D}_{\texttt{place}}\}$, we then construct the sequence of manipulation tasks, $\mathcal{M} = \langle (\texttt{pick}, \mathcal{O}'_{\texttt{pick}}), (\texttt{pour}, \mathcal{O}'_{\texttt{pour}}),  (\texttt{place}, \mathcal{O}'_{\texttt{place}})\rangle$, for the robot to perform. The intended outcome of the sequence $\mathcal{M}$ is for the robot to pick a beaker placed at $\mathcal{O}'_{\texttt{pick}}$, pour its contents into a round bottom flask located at $\mathcal{O}'_{\texttt{pour}}$ and finally place the beaker at $\mathcal{O}'_{\texttt{place}}$. To validate the planner, we randomly varied the poses, $\mathcal{O}'_{\texttt{pick}}, \mathcal{O}'_{\texttt{pour}}, \mathcal{O}'_{\texttt{place}}$, across each trial, to test its ability to generalize to new task instances. We conducted a total of 10 trials of this $\texttt{pick}-\texttt{pour}-\texttt{place}$ manipulation sequence by executing it on the robot. The robot was able to successfully pick the beaker, pour its contents into the flask and place it for all the conducted 10 trials. We have included a supplementary video documenting all the conducted trials here \texttt{\url{https://youtu.be/8B_9PeFMtKI}}.}

\subsection{Case Study on Nanoparticle Synthesis}

{To highlight the translational aspect of our approach to synthesize chemicals, we conducted two case studies on the synthesis of nanoparticles using robots. We collected a single kinesthetic demonstration for each of the manipulation task involved in the synthesis. The user then specified the sequence of tasks, $\mathcal{M}$, that need to be carried out in order for the solution to be synthesized. Using the User Guided Planner, we then executed the sequence $\mathcal{M}$ on the robot to synthesize the nanoparticles and then characterized a sample of the synthesized solution. Both case studies were run once.}
For more details about the hardware setup and the conducted experiments, please refer to the following video \texttt{\url{https://youtu.be/gBd9wzv8Cgs}}.

\subsubsection{Gold Nanoparticle Synthesis}
The facile synthesis of Au NPs~\cite{premkumar2007goldnp} was selected as a preliminary test to evaluate the capability of the system for nanoparticle synthesis. 
Two reaction conditions were run simultaneously by the robot: one without stirring and the other with stirring. The reduction of potassium chloroaurate ($\mathrm{KAuCl_4}$) by a polysorbate 80 (T-80), a non-ionic surfactant, can be completed by simple mixing. 
Stock solutions of 1 weight percent polysorbate 80 in water and 0.4 mM $\mathrm{KAuCl_4}$ were prepared and provided in reservoirs connected to solenoid controlled liquid dispensers. 
The robot picked up a nearby scintillation vial, moved it to each liquid dispenser, and collected 10 mL of polysorbate 80 and 4 mL of $\mathrm{KAuCl_4}$. The robot subsequently moved the vial onto a nearby stand. To ensure that the sample was not exposed to light for the duration of the experiment, it lifted a provided opaque plastic lid and placed it to cover the vial. The robot then collected the same volume of each reagent in a second scintillation vial that had been prepared to include a magnetic stir bar. The robot placed this vial on the stir plate, covered it with a similar opaque plastic lid, and turned the knob on the plate to begin stirring. The solution was stirred at moderate speed for the duration of the experiment.  After one hour had passed, the robot lifted each of the plastic lids to reveal the vials, allowing the camera on the robot to photograph each of the sample vials. The lids were subsequently replaced. Photography was repeated every two hours, following the progression of this process until a total of five hours had elapsed.  Following each instance of visual photographic assessment, 
an aliquot of the sample was collected by a human chemist for characterization via transmission electron microscopy. 

\begin{figure}[ht!]
    \centering
    \setlength{\tabcolsep}{1pt}
    
    \begin{tabular}{c c c c}
        
        & 1 Hour & 3 Hours & 5 Hours \\
        
        \rotatebox[origin=c]{90}{Stirring} &
        \begin{subfigure}[c]{0.28\columnwidth}
            \includegraphics[width=\linewidth]{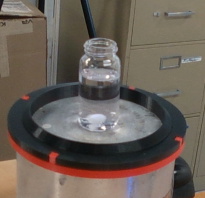}
        \end{subfigure} &
        \begin{subfigure}[c]{0.28\columnwidth}
            \includegraphics[width=\linewidth]{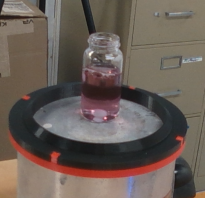}
        \end{subfigure} &
        \begin{subfigure}[c]{0.28\columnwidth}
            \includegraphics[width=\linewidth]{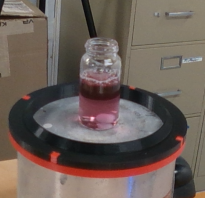}
        \end{subfigure} \\

        \noalign{\vspace{2pt}}
        
        \rotatebox[origin=c]{90}{Not Stirring} &
        \begin{subfigure}[c]{0.28\columnwidth}
            \includegraphics[width=\linewidth]{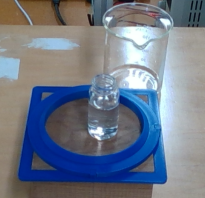}
        \end{subfigure} &
        \begin{subfigure}[c]{0.28\columnwidth}
            \includegraphics[width=\linewidth]{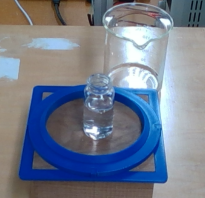}
        \end{subfigure} &
        \begin{subfigure}[c]{0.28\columnwidth}
            \includegraphics[width=\linewidth]{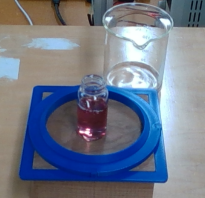}
        \end{subfigure}%
        
    \end{tabular}
    
    \caption{Observable differences in gold nanoparticle concentration when the robot was instructed to perform simple lab tasks, such as stirring.}
    \label{fig:gold_np_collected_images}
\end{figure}

\begin{figure}[ht!]
    \centering
    \includegraphics[width=0.95\linewidth]{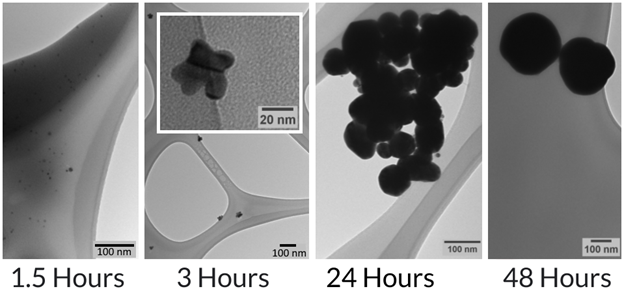}
    \caption{Gold Nanoparticle growth over time.}
    \label{fig:gold_np_tem}
\end{figure}

\begin{figure}[ht!]
    \centering
    \setlength{\tabcolsep}{1pt}
    
    \begin{tabular}{c c c c}
        
        & 1 Hour & 3 Hours & 5 Hours \\
        
        \rotatebox[origin=c]{90}{Stirring} &
        \begin{subfigure}[c]{0.28\columnwidth}
            \includegraphics[width=\linewidth]{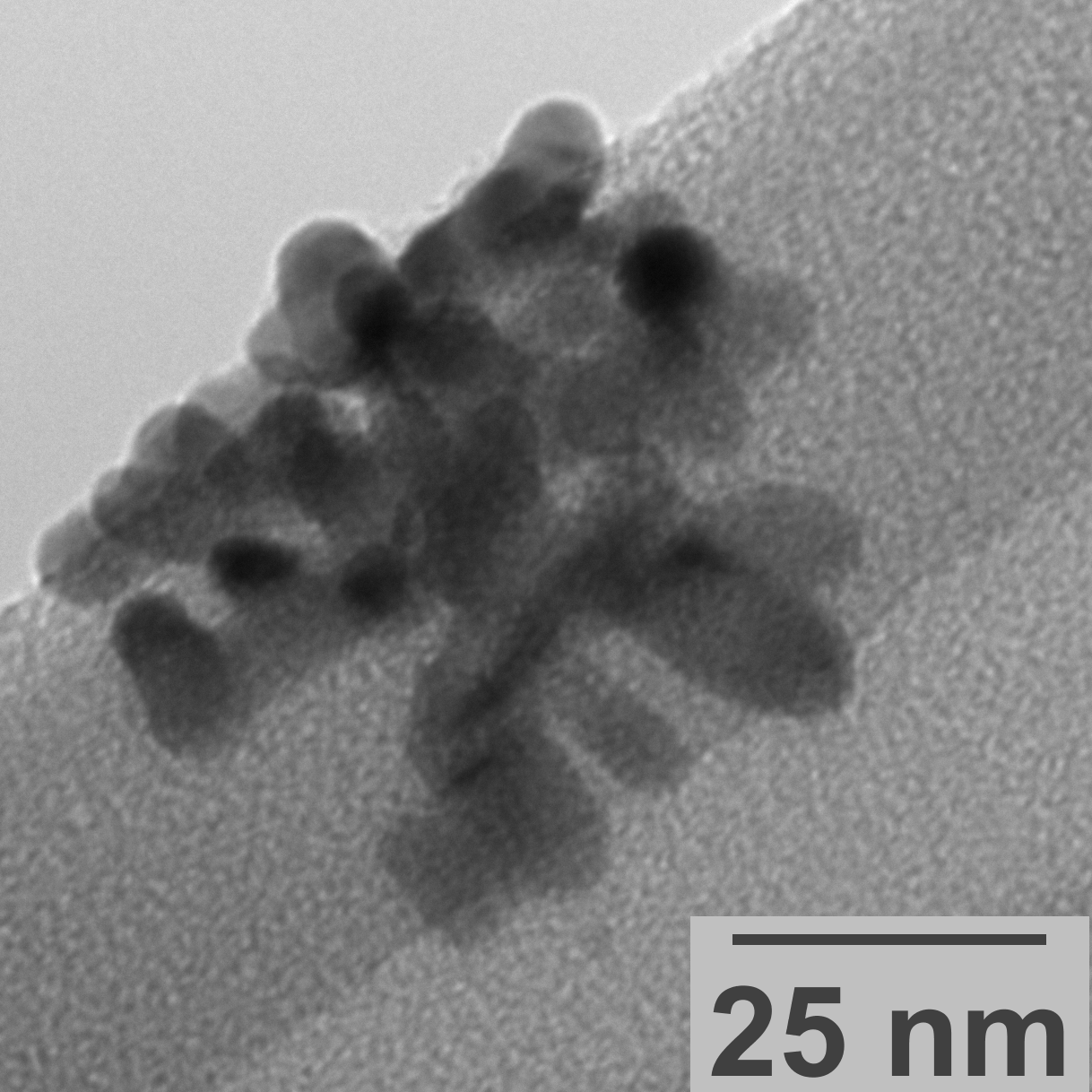}
        \end{subfigure} &
        \begin{subfigure}[c]{0.28\columnwidth}
            \includegraphics[width=\linewidth]{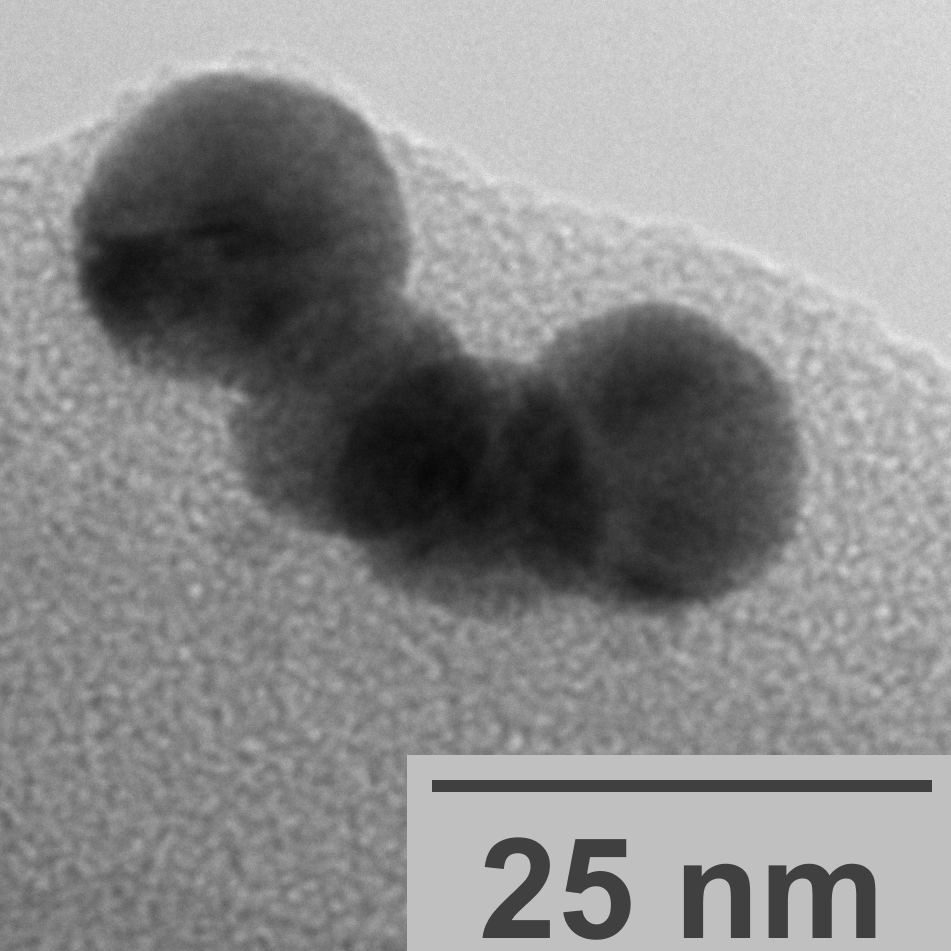}
        \end{subfigure} &
        \begin{subfigure}[c]{0.28\columnwidth}
            \includegraphics[width=\linewidth]{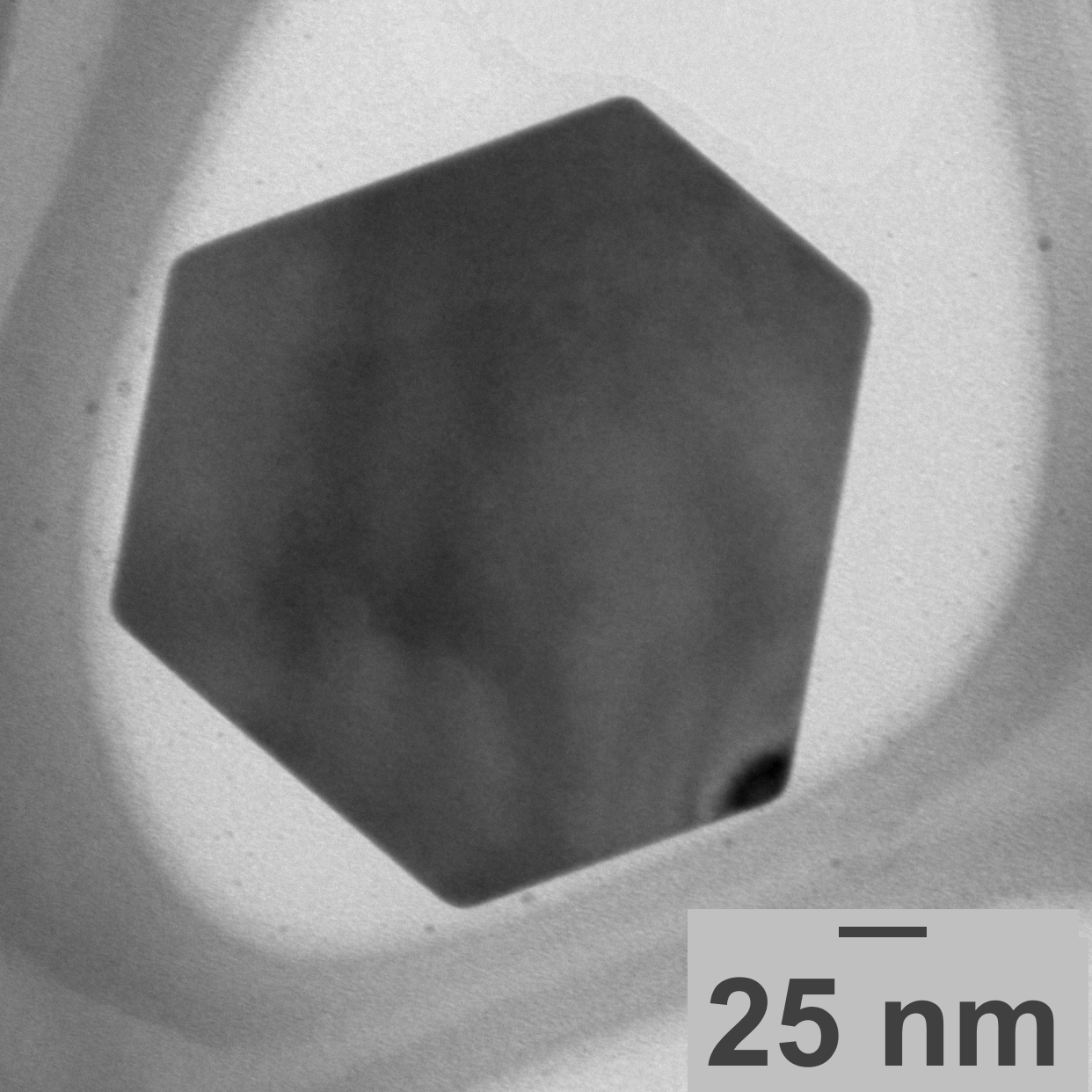}
        \end{subfigure} \\

        \noalign{\vspace{2pt}}
        
        \rotatebox[origin=c]{90}{Not Stirring} &
        \begin{subfigure}[c]{0.28\columnwidth}
            \includegraphics[width=\linewidth]{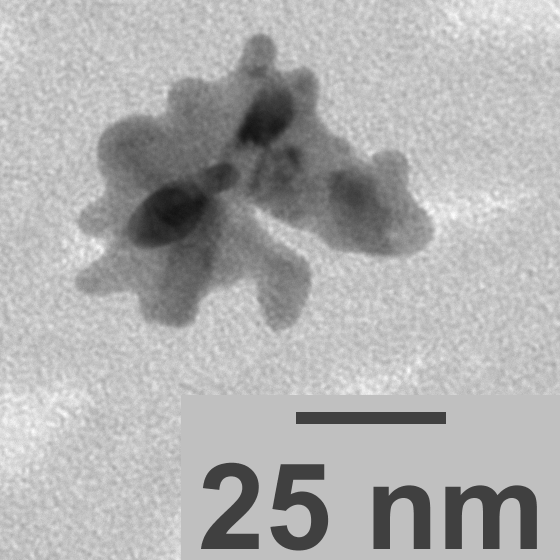}
        \end{subfigure} &
        \begin{subfigure}[c]{0.28\columnwidth}
            \includegraphics[width=\linewidth]{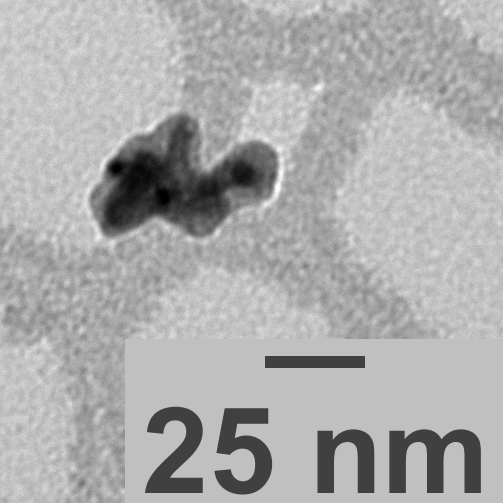}
        \end{subfigure} &
        \begin{subfigure}[c]{0.28\columnwidth}
            \includegraphics[width=\linewidth]{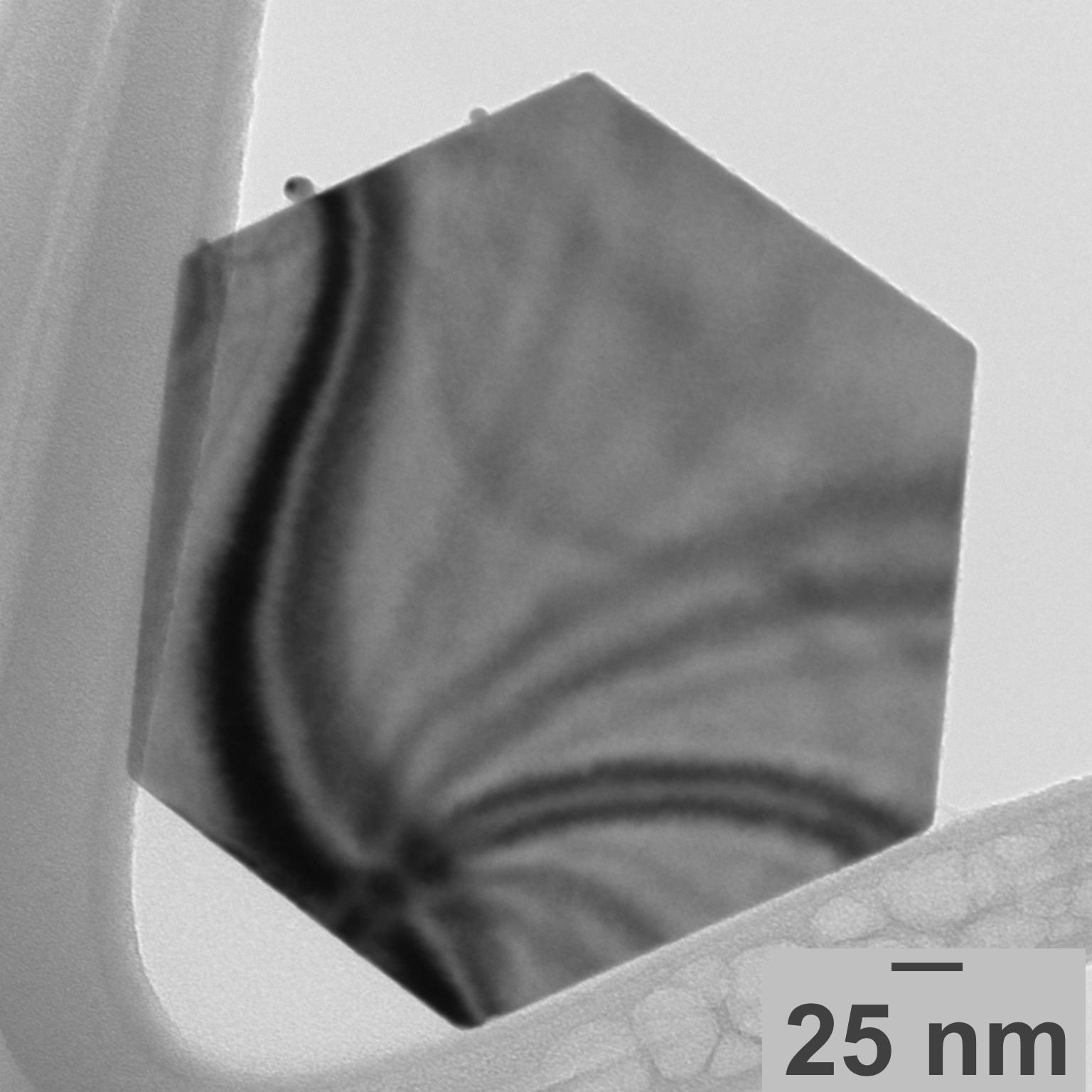}
        \end{subfigure}%
        
    \end{tabular}
    
    \caption{Comparison of as-synthesized gold nanoparticles, with and without stirring.}
    \label{fig:gold_np_stirring_series_tem}
\end{figure}

\begin{figure*}[ht!]
    \begin{subfigure}[b]{\textwidth}
        \includegraphics[width=\textwidth]{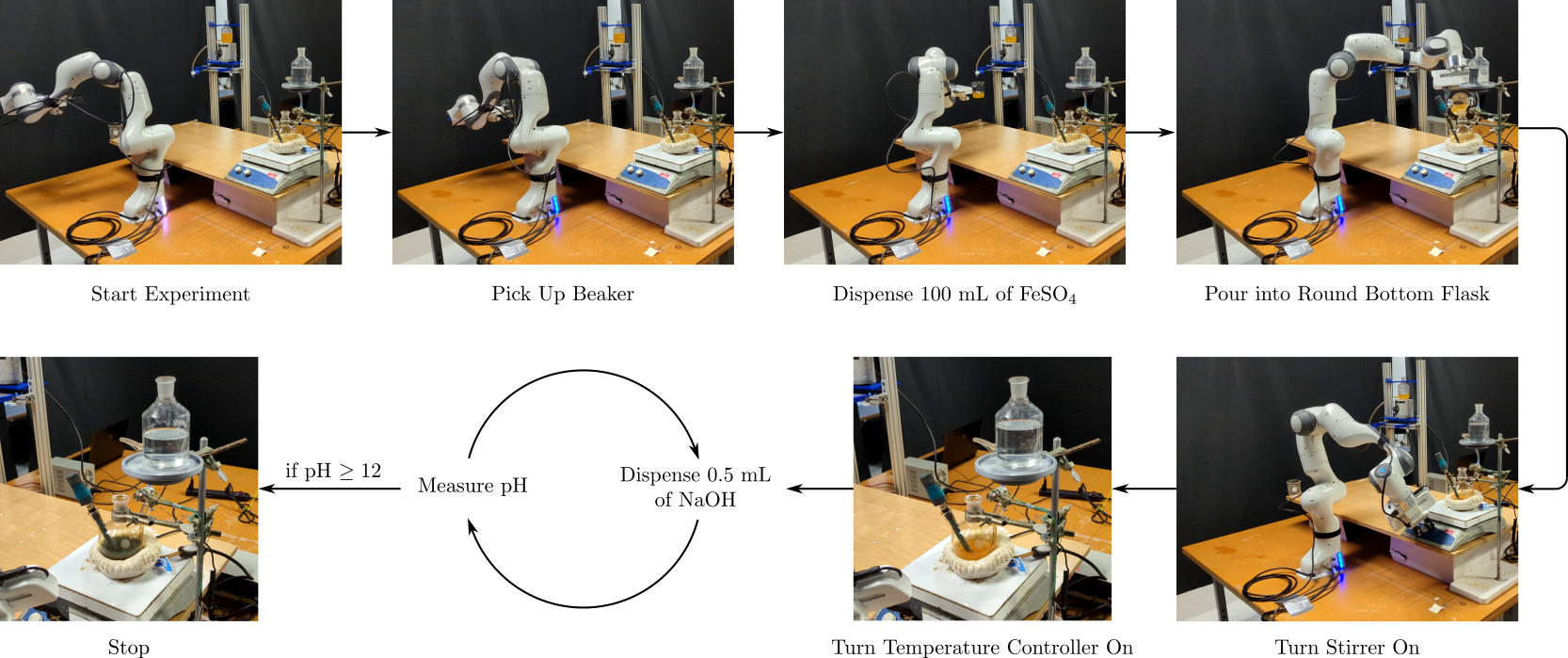}
    \end{subfigure}
    \caption{{Automated Magnetite Nanoparticle Synthesis using Robots:} Sequence of tasks involved in the synthesis of Magnetite Nanoparticles.}
    \label{fig:magnetite_np_exp}
\end{figure*}

\vspace{0.5\baselineskip}
\noindent
\textit{\textbf{Task Setup:}}
For this experiment, the user provided a single kinesthetic demonstration for each of the following manipulation tasks: 1) Pick up object ($\mathcal{D}_{\texttt{pick}}$), 2) Place object ($\mathcal{D}_{\texttt{place}}$), 3) Move to dispenser ($\mathcal{D}_{\texttt{predispense}}$), 4) Move away from dispenser ($\mathcal{D}_{\texttt{postdispense}}$),
5) Turn on stirrer ($\mathcal{D}_{\texttt{stirrer}}$), and 6) Move the end effector to view the scintillation vial ($\mathcal{D}_{\texttt{view}}$). For dispensing the solution and placing the vial, the following sequence of tasks was used $\langle (\texttt{pick}, \mathcal{O}_{\texttt{vpick}}), ~(\texttt{predispense}, \mathcal{O}_{\mathrm{KAuCl_4}}), \allowbreak (\texttt{postdispense}, \mathcal{O}_{\mathrm{KAuCl_4}}), ~(\texttt{predispense},\mathcal{O}_{\mathrm{T-80}}), \allowbreak (\texttt{postdispense},\mathcal{O}_{\mathrm{T-80}}), (\texttt{place}, \mathcal{O}_{\texttt{vplace}}) \rangle$ where $\mathcal{O}_{\texttt{vpick}}$ and $\mathcal{O}_{\texttt{vplace}}$ correspond to the pose at which to pick the scintillation vial from and the pose at which to place the scintillation vial after dispensing the solutions respectively. To pick and close the vials with the opaque plastic lid, we used the following sequence of tasks, $\langle (\texttt{pick}, \mathcal{O}_{\texttt{cpick}}), ~(\texttt{place}, \mathcal{O}_{\texttt{cplacei}}), ~(\texttt{pick}, \mathcal{O}_{\texttt{cplacei}}), \allowbreak (\texttt{place}, \mathcal{O}_{\texttt{cplace}}) \rangle$. Please note that for picking and placing the cover, we first pick the cover from $\mathcal{O}_{\texttt{cpick}}$, place the cover at an intermediate pose $\mathcal{O}_{\texttt{cplacei}}$, and then pick the cover from the intermediate pose with a different grasp to finally place the cover over the scintillation vial. This was done to prevent motion plan failure, as the required motion when using a single grasp caused the manipulator to go outside its workspace. Please refer to the supplementary video for more details. 

To turn on the stirrer, we perform the manipulation task $\langle (\texttt{stirrer}, \mathcal{O}_{\texttt{stirrer}}) \rangle$ where $\mathcal{O}_{\texttt{stirrer}}$ is the pose of the stir plate. To check the progress of the reaction, we need to remove the cover, move the end-effector to view the scintillation vial, and then replace the cover. We use the sequence of manipulation tasks $\langle (\texttt{pick}, \mathcal{O}_{\texttt{cplace}}), ~(\texttt{place}, \mathcal{O}_{\texttt{cplacei}}), ~(\texttt{pick}, \mathcal{O}_{\texttt{cplacei}}), \allowbreak (\texttt{view}, \mathcal{O}_{\texttt{vplace}}), ~(\texttt{place}, \mathcal{O}_{\texttt{cplace}}) \rangle$. We pick the cover from where it was finally placed in the previous step, place it in an intermediate configuration, move the end-effector to bring the scintillation vial within the field of view of the camera, capture an image and then replace the cover on the vial. We repeat this process in a loop until the given time duration has passed with a time delay between each iteration. In our experiment, we checked the reaction progress at 1 hour, 3 hours and 5 hours since the experiment starts. The images captured by the robot are shown in Figure \ref{fig:gold_np_collected_images}.
For segmenting the vial in the image, we use SAM \cite{kirillov2023segany}. The prompt given by the user as a mouse click on the RGB image captured by the camera was used to segment the vial containing the solution for determining the solution color.
Figure \ref{fig:gold_np_exp} shows the sequence of steps that were performed in this experiment. Using the provided sequence of manipulation tasks, the robot autonomously determined the joint space path that the robot needs to go through to perform the manipulation tasks and execute them in real-time.

{Please note that for this experiment, for the \texttt{pick} and \texttt{pour} manipulation tasks, we collected a single demonstration for each, $\mathcal{D}_{\texttt{pick}}$ and $\mathcal{D}_{\texttt{place}}$ respectively. While we used a scintillation vial to provide the demonstration, we were also able to transfer this demonstration for performing the \texttt{pick} and \texttt{place} tasks on the beaker and the scintillation vial cover. We also changed the object poses compared to the poses during the demonstration as well as the grasp poses to highlight the capabilities of the User Guided Motion Planner.}

\vspace{0.5\baselineskip}
\noindent
\textit{\textbf{Structural and Chemical Characterization:}}
Three series of gold nanoparticles were completed as part of these demonstrations. Each of these runs probed a different parameter controllable by the robot. Initial tests, before stirring was implemented, showed little influence on particle size as a result of varying concentration. By contrast, particle sizes increased with longer reaction times, as one might expect from a reaction that is limited by sample diffusion and the low reduction potential of polysorbate 80. Specifically, longer reaction times enabled the growth of larger particles as incoming gold ions in solutions were slowly coordinated by polysorbate 80 onto the surfaces of the growing nanoclusters and were consequently reduced. These findings are confirmed using TEM, as shown in Figure \ref{fig:gold_np_tem}. Notably, particles synthesized in this way were similar to those presented in the work being replicated~\cite{premkumar2007goldnp}. From these findings, we can conclude that the slow reduction of the $\mathrm{Au^{3+}}$ ions is mediated entirely by the availability of polysorbate 80 ions to serve as reducing agents. When trained to operate a stir plate, the robot was able to demonstrate a clear improvement in the rate of reaction. Specifically, an earlier color change was indicative of a faster reduction of Au NPs, as noted in Figure \ref{fig:gold_np_collected_images}. By comparison, while the evolution of larger particles was (not surprisingly) observed first in the stirred sample, these particles did not show any significant changes in size and morphology after five hours, as shown in Figure \ref{fig:gold_np_stirring_series_tem}.

By being able to replicate the gold nanoparticle synthesis protocol with a robotic setup, we enable the exploration of various variants of the experiments, which is difficult to conduct by humans because of the enormity of the search space. For example, future experiments may involve a number of strategies including the heating of the reaction vessel, careful control of the stirring rate, and/or increasing concentrations of polysorbate 80.

\subsubsection{Magnetite Nanoparticle Synthesis}

Following the successful reproduction of the Au NP synthesis procedure, to highlight the versatility of our approach, we selected a different protocol to synthesize magnetite nanoparticles, where the robot also has to use the skill of pouring.

The reaction followed a facile 1-step protocol \cite{suppiah2016magnetitenp}, in which $100$ mL of $0.45$ M $\mathrm{FeSO_4}$ was oxidized to magnetite via the addition of $0.45$ M NaOH.  Solutions of reagents were prepared by human operators, and then supplied to the robot using the same calibrated liquid dispensers discussed before.  Due to the low solubility of $\mathrm{FeSO_4}$ in water, the reservoir containing this reagent was left to stir for the duration of the reaction. The robot was subsequently provided with a $250$ mL beaker and a $500$ mL round-bottom flask mounted in a heating mantle and above a stir plate.  A temperature probe and pH probe were inserted into the round bottom flask, prior to experimentation. 

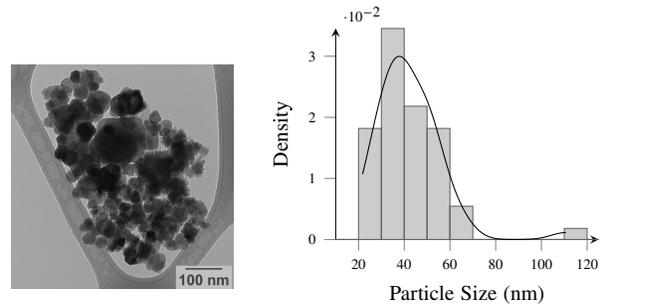
\begin{figure}[ht!]
    \centering

    \begin{subfigure}[b]{0.35\linewidth}
        \includegraphics[width=\linewidth]{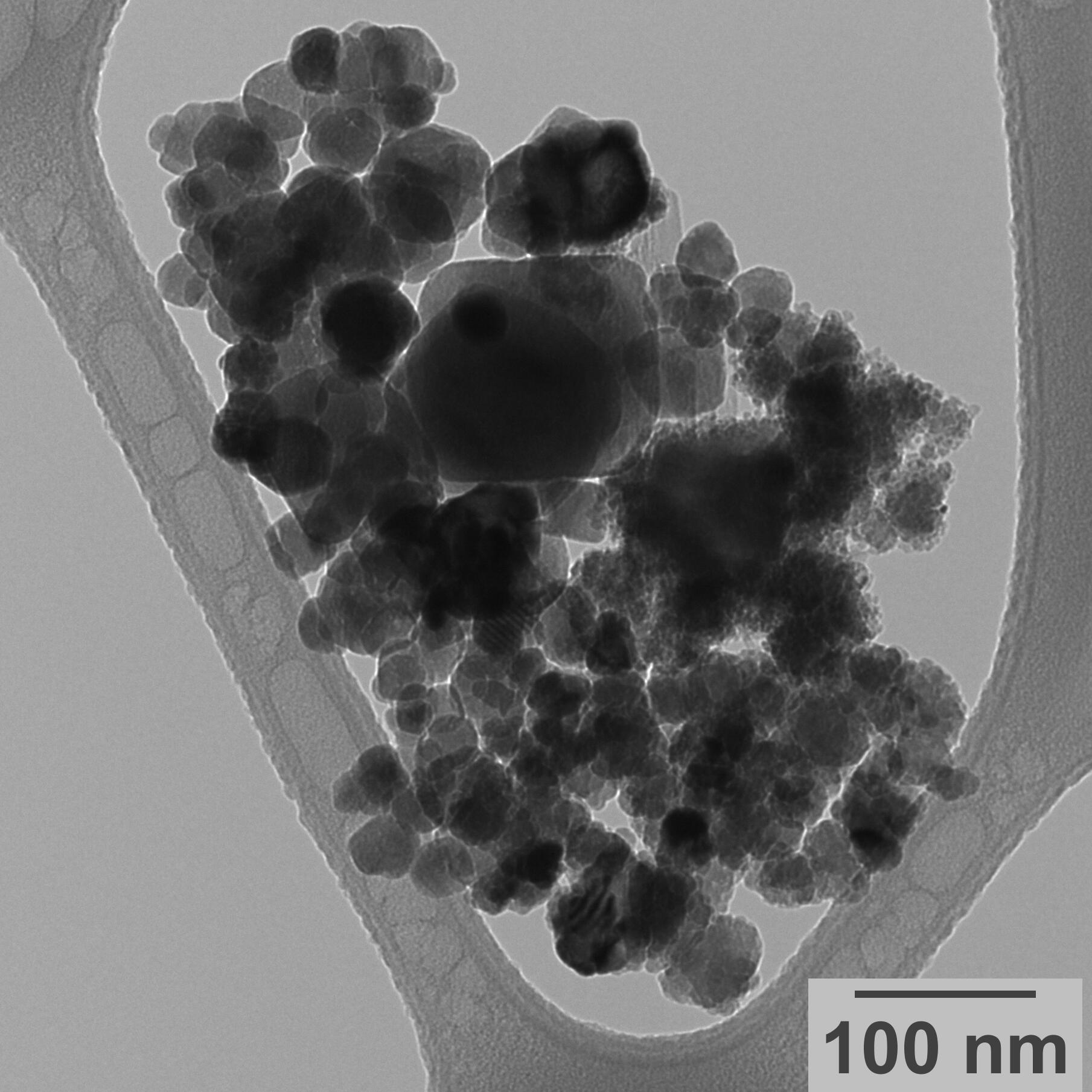}
        \caption{TEM image of magnetite nanoparticles, as synthesized by the robot.}
    \end{subfigure}
    \hfill
    \begin{subfigure}[b]{0.6\linewidth}
    
        \begin{tikzpicture}
            \begin{axis}[
                width=\linewidth,
                xlabel={Particle Size (nm)},
                ylabel={Density},
                xmin=10, xmax=125,
                axis lines=left,
                tick align=outside,
                xtick={20, 40, 60, 80, 100, 120},
                tick label style={font=\tiny},
                label style={font=\small}
            ]
                
                \addplot[
                    ybar interval,
                    fill=black!20,  
                    draw=black!60,  
                    hist={
                        bins=13,
                        data min=0,
                        data max=130,
                        density=true
                    }
                ] table [
                    col sep=comma, 
                    y=Length,
                ] {figures/data/magnetite_run_3_particle_size.csv};
    
                 \addplot[
                    color=black,
                ]
                table[
                    col sep=comma,
                    x=x,
                    y=density
                ]{figures/data/magnetite_run_3_particle_size_density.csv};
                
            \end{axis}
        \end{tikzpicture}

        \caption{Size distribution of isolated, analyzed particles, measured over 50 particles.}
    \end{subfigure}
    
    \caption{Characterization of magnetite nanoparticles synthesized by the robot.}
    \label{fig:magnetite_np_tem}
\end{figure}

In sequence, the robot picked up the beaker, navigated to the $\mathrm{FeSO_4}$ dispenser, collected 100 mL, and delivered it into the round-bottom flask.  Following these steps, the robot powered on the stir plate, and the temperature controller was also turned on.  When 45 °C was measured, the robot began adding in NaOH dropwise.  After each series of drops were added in, the associated pH values were measured again and recorded.  The number of drops added varied as the volume in the reservoir decreased, in order to maintain the addition of a level of constant volume within the reaction vessels.  When a pH of $12$ was achieved, the reaction was completed.  The temperature controller and stir plate were switched off, and the reaction vessel was allowed to cool before being collected for washing and characterization by a human chemist. The resulting product consisted of a black powder.

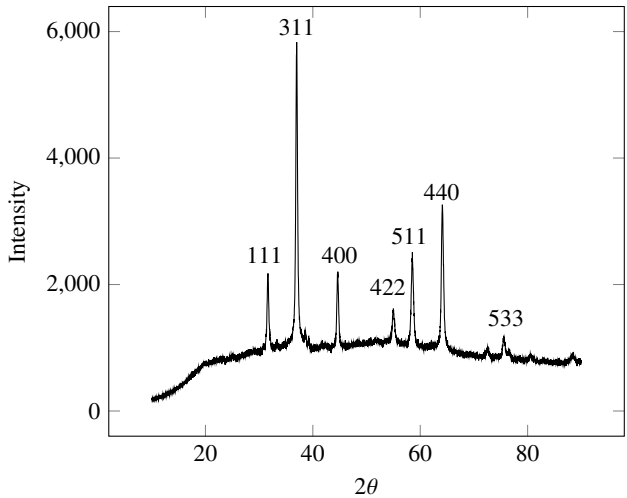
\begin{figure}[ht!]
    \centering
    
    \begin{tikzpicture}
    \begin{axis}[
        xlabel={$2\theta$},
        ylabel={Intensity},
    ]
    
    \addplot[
        color=black,
    ]
    table[
        col sep=comma,
        x=two_theta,
        y=intensity
    ]{figures/data/magnetite_run_3_xrd.csv};
    
    \node[above] at (axis cs:31,2200) {111};
    \node[above] at (axis cs:37,5800) {311};
    \node[above] at (axis cs:45,2200) {400};
    \node[above] at (axis cs:54,1700) {422};
    \node[above] at (axis cs:58,2500) {511};
    \node[above] at (axis cs:64,3200) {440};
    \node[above] at (axis cs:76,1200) {533};
    
    \end{axis}
    \end{tikzpicture}
    
    \caption{XRD pattern of magnetite nanoparticles synthesized by the robot. Major peaks associated with magnetite are labeled. }
    \label{fig:magnetite_np_xrd}
\end{figure}

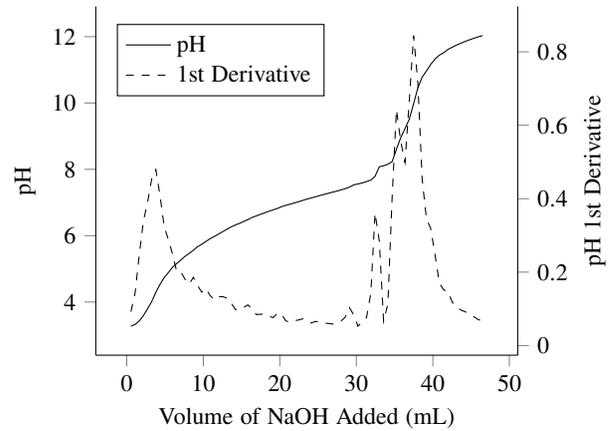
\begin{figure}
    \centering
    
    \begin{tikzpicture}
    
    \begin{axis}[
        width=0.85\linewidth,
        axis y line*=left,
        axis x line*=bottom,
        legend semaphore/.style={draw=none},
        enlargelimits=0.1,
        xlabel={Volume of NaOH Added (mL)},
        ylabel={pH},
        tick align=outside,
        ytick={4, 6, 8, 10, 12},
    ]
    
    \addplot[
        color=black,
    ]
    table[
        col sep=comma,
        x=total_dispensed_ml,
        y=measured_pH
    ]{figures/data/magnetite_run_3_titration.csv};
    \label{tz:ph_plot}
    \end{axis}

    \begin{axis}[
        width=0.85\linewidth,
        axis y line*=right,
        axis x line=none,
        enlargelimits=0.1,
        ylabel={pH 1st Derivative},
        tick align=outside,
        legend style={
          at={(0.05,0.95)}, 
          anchor=north west,
          cells={anchor=west}
        },
    ]
    \addlegendimage{/pgfplots/refstyle=tz:ph_plot}\addlegendentry{pH}
    \addplot[
        color=black, dashed
    ]
    table[
        col sep=comma,
        x=total_dispensed_ml,
        y=pH_first_derivative
    ]{figures/data/magnetite_run_3_titration.csv};
    \addlegendentry{1st Derivative}
    \end{axis}
    
    \end{tikzpicture}
    
    \caption{Magnetite titration curve and corresponding first derivative.}
    \label{fig:magnetite_np_titration_curve}
\end{figure}

\vspace{0.5\baselineskip}
\noindent
\textit{\textbf{Task Setup:}}
For this experiment, the user provided a single kinesthetic demonstration for each of the following manipulation tasks: 1) Turn on stirrer ($\mathcal{D}_{\texttt{stirrer}}$) 2) Pouring solution from container ($\mathcal{D}_{\texttt{pour}}$). We also reused the demonstrations $\mathcal{D}_{\texttt{pick}}$, $\mathcal{D}_{\texttt{place}}$, $\mathcal{D}_{\texttt{predispense}}$
from the previous task. This shows that we can create a reusable database of manipulation primitives that can be used across different experiments. The only reason we had to provide a new demonstration for stirring is because we wanted to use a different speed setting for the stir plate which meant that the control knob had to be rotated to a different angular position compared to the previous experiment. The sequence of manipulation tasks involved in this experiment are $\langle (\texttt{pick}, \mathcal{O}_{\texttt{bpick}}), ~(\texttt{predispense}, \mathcal{O}_{\mathrm{FeSO_4}}), \allowbreak 
(\texttt{pour}, \mathcal{O}_{\texttt{flask}}), \allowbreak (\texttt{place}, \mathcal{O}_{\texttt{bppick}}) \rangle$. The robot picks a beaker, dispenses $100$ mL of $\mathrm{FeSO_4}$ and then pours it into a round bottom flask. It then places the beaker at the same pose from where it picked it up and then turns the stir plate on. After this sequence of manipulation tasks is completed, the Arduino turns on the heating mantle and then runs the titration process by dispensing $4$ mL of $\mathrm{NaOH}$, measuring the pH of the solution and repeating this process until a pH of $12$ was achieved. The titration curve of the reaction is shown in Figure \ref{fig:magnetite_np_titration_curve}. Figure \ref{fig:magnetite_np_exp} shows the sequence of steps that were performed to synthesize Magnetite. Similar to the previous experiment, the robot used the provided sequence of manipulation tasks to determine the joint space path to perform the given manipulation tasks and to execute it in real-time.

\vspace{0.5\baselineskip}
\noindent
\textit{\textbf{Structural and Chemical Characterization:}}
Magnetite particles were imaged by both TEM and X-ray diffraction (XRD) analysis, as highlighted in Figures \ref{fig:magnetite_np_tem} and \ref{fig:magnetite_np_xrd}. TEM analysis of these particles yields a morphology that closely resembles spheres similar to those presented in the referenced work, with the isolated particles showing the expected size of 42.3 ± 14.2 nm. We further investigated the crystal structure of the as-synthesized particles by XRD analysis, which produces a series of characteristic peaks resulting from the arrangement of atoms within repeating crystal layers. These layers have predicted lengths that result from the different bond lengths between atoms. Consequently, XRD analysis allows for a quantitative assessment of the crystal make-up and a qualitative assessment of the crystallinity and purity of a sample. In this case, XRD analysis gave rise to a pattern similar to that observed in the original work, with a reduced $\alpha$-FeOOH impurity, especially when compared with the previously published studies. This pattern clearly evinced the formation of magnetite.

Finally, upon reviewing the pH output readings recorded by the robot during the reaction, as seen in Figure \ref{fig:magnetite_np_titration_curve}, we found a similar pattern analogous to that of the published findings. Notably, two clear inflection points are visible at pH 4.2 and 10.4. These inflection points indicate regions in which the concentration of sodium hydroxide in solution is high enough for a change in the equilibrium state of the dissolved iron. The first inflection point, which occurs immediately as sodium hydroxide is added to the reaction mixture, is the hydrolysis of iron sulfate to goethite (FeOOH), an iron hydroxide intermediate. As more sodium hydroxide is added, the pH steadily increases until the pH is high enough to form iron (II) hydroxides, which promptly react with the as-formed goethite, resulting in the formation of magnetite ($\mathrm{Fe_2O_3}$) above pH 10.4. Both inflection points were detected at nearly the same pH as in the published work, thereby giving credence to the suitability and accuracy of our system. It is worth noting that the peaks for the first derivative are wider than one would typically expect. This observation was the result of our current titrator delivering solutions dropwise, rather than as a continuous flow. Modules that deliver a solution continuously are commercially available and may be added to our system in the future, enabling the collection of a higher quality and more precise set of data.

\section{Conclusion}
\label{sec:conc}

In this work, we have proposed a screw geometry based manipulation planning framework that allows us to build a library of reusable manipulation skills using kinesthetic demonstrations, which can then be composed to perform multi-step laboratory protocols. The proposed framework can extract the complex motion constraints that are implicit in the provided demonstration and ensure that the generated motion plans can satisfy such constraints. To show the effectiveness of our framework, we conducted solution-based synthesis of nanoparticles which involves a sequence of manipulation tasks with motion constraints. We provided the kinesthetic demonstrations to build a library of reusable manipulation skills, and put these together to create the multi-step protocol required for the chemical synthesis. We then used the ScLERP based planner combined with Jacobian pseudoinverse to execute the tasks with actual reagents for synthesizing nanoparticles in a laboratory setting. We also showed that the manipulation skills can be transferred from one chemical synthesis protocol to another, proving that the manipulation skills can be reused. We characterized the structural and chemical properties of the nanoparticles that were synthesized using our framework to evaluate the effectiveness of our framework. The proposed framework is general enough to be extended to other chemical protocols involving constrained manipulation.

{Please note that while we did not observe any failures in the conducted experiments, in general the User Guided Motion Planner does not generate successful motion plans for all possible task-instances. Failures may occur due manipulator joint position limit violation, manipulator singularities and in some cases the required motion falling out of the manipulator workspace. Overcoming these issues have been studied in \cite{das2026bandit} where multiple demonstrations are systematically collected for each manipulation task to ensure that the robot has a probabilistic confidence bound in its ability to generate feasible manipulation plans within a predefined work area.  If the robot cannot perform a manipulation task due to some change in the placement of objects, as~\cite{das2026bandit} shows, it can actually compute beforehand that it cannot do the task and then ask for another example. However, we do not explore this aspect here, and it will be part of future work.}

While the proposed framework indicates considerable promise in performing a fixed protocol for the synthesis, it can be extended to vary parameters within the multi-step protocol to potentially explore the automated synthesis of chemicals. Moreover, the current implementation does not use sensor information to perceive the environment for detecting and localizing the chemical apparatus and other equipment. Future work would involve extending this framework to incorporate a structured approach to vary protocol parameters and sensing.

\section*{Funding Data}

\begin{itemize}
\item Experiments were financially supported in part by seed grant funding through the Office of the Vice President for Research at Stony Brook University.
\item Structural characterization experiments (TEM and SEM) were performed in part at the Center for Functional Nanomaterials, located at Brookhaven National Laboratory, which is supported by the U.S. Department of Energy under Contract No. DE-SC0012704.
\end{itemize}


\bibliographystyle{asmejour}

\bibliography{references}


\end{document}